\documentclass[a4paper,11pt]{article}

\usepackage[margin=1.in]{geometry}
\usepackage{cite}
\usepackage[pdftex]{graphicx}
\usepackage[utf8]{inputenc} 
\usepackage[T1]{fontenc}    
\usepackage{lmodern}
\usepackage{hyperref}       
\hypersetup{colorlinks,linkcolor=blue, citecolor=blue}
\usepackage{url}            
\usepackage{booktabs}       
\usepackage{amsfonts}       
\usepackage{nicefrac}       
\usepackage{microtype}      
\usepackage{multicol}
\usepackage{xcolor}
\usepackage{multicol}

\usepackage{amsmath}
\usepackage{amssymb}
\usepackage{bm}
\usepackage{float}
\usepackage{graphicx}
\usepackage{caption}
\usepackage{subcaption}
\usepackage{comment}

\usepackage{titlesec}      
\usepackage{titlesec,titletoc} 

\usepackage[hang,flushmargin]{footmisc}

\def\<{\langle} \def\>{\rangle}

\DeclareMathOperator*{\argmax}{arg\,max}
\DeclareMathOperator*{\argmin}{arg\,min}

\usepackage{amsthm}
\newtheorem{theorem}{Theorem}[section]
\newtheorem{lemma}[theorem]{Lemma}

\title{The twin peaks of learning neural networks}

\author{
	Elizaveta Demyanenko$^1$, Christoph Feinauer$^1$, Enrico M. Malatesta$^1$, Luca Saglietti$^1$
}
\date{$^1$ \textit{Department of Computing Sciences, Bocconi University, 20136 Milano, Italy}}

\usepackage[toc,page]{appendix}

\renewcommand{\appendixtocname}{}

\makeatletter
\let\oldappendix\appendices

\renewcommand{\appendices}{%
	\clearpage
	\renewcommand{\thesection}{\Roman{section}}
	\let\tf@toc\tf@app
	\addtocontents{app}{\protect\setcounter{tocdepth}{5}}
	\immediate\write\@auxout{%
		\string\let\string\tf@toc\string\tf@app^^J
	}
	\oldappendix
}%

\newcommand{\listofappendices}{%
	\begingroup
	\renewcommand{\contentsname}{\appendixtocname}
	\let\@oldstarttoc\@starttoc
	\def\@starttoc##1{\@oldstarttoc{app}}
	\tableofcontents
	\endgroup
}

\makeatother
\usepackage{imakeidx}
\makeindex

\begin{document}

	\maketitle
	
	\tableofcontents
	\newpage
	\begin{abstract}
		Recent works demonstrated the existence of a double-descent phenomenon for the generalization error of neural networks, 
		where highly overparameterized models escape overfitting and achieve good test performance, at odds with the standard bias-variance trade-off described by statistical learning theory. 
		In the present work, we explore a link between 
		this phenomenon and the increase of complexity and sensitivity of the function represented by neural networks. In particular, we study the Boolean mean dimension (BMD), a metric developed in the context of Boolean function analysis.
		Focusing on a simple teacher-student setting for the random feature model, we derive a theoretical analysis based on the replica method that yields an interpretable expression for the BMD, in the high dimensional regime where the number of data points, the number of features, and the input size grow to infinity.
		We find that, as the degree of overparameterization of the network is increased, the BMD reaches an evident peak at the interpolation threshold, in correspondence with the generalization error peak, and then slowly approaches a low asymptotic value. The same phenomenology is then traced in numerical experiments with different model classes and training setups. 
		Moreover, we find empirically that adversarially initialized models tend to show higher BMD values, and that models that are more robust to adversarial attacks exhibit a lower BMD.
	\end{abstract}

	
	\section{Introduction}
	
	The evergrowing scale of modern neural networks often prevents a detailed understanding of how predictions relate back to the model inputs \cite{sejnowski2020unreasonable}. While this lack of interpretability can hinder adoption in sectors with a high impact on society \cite{rudin2019stop}, the impressive performance of neural network-based models in fields like natural language processing \cite{vaswani2017attention, openai2023gpt4, touvron2023llama}, computational biology \cite{jumper2021highly} and computer vision and image generation \cite{ramesh2022hierarchical, rombach2022high} have made them the de-facto standard for many real-world applications. This tension has motivated a large interest in the field of explainable AI (XAI) \cite{montavon2018methods, guidotti2018survey, vilone2020explainable}.
	
	Deep learning models, which by now can feature hundreds of billions of parameters \cite{brown2020language}, seemingly defy the notion that increasing model complexity should decrease generalization performance. Counter to what one would expect from statistical learning theory \cite{vapnik1999overview}, the observation has been that larger ---heavily overparameterized--- models often perform better \cite{neyshabur2017exploring}. This has led to the question how complex the function represented by an overparameterized neural network is after training. Many lines of research suggest that neural network models are biased towards implementing simple functions, despite their large parameter count, and that this implicit bias is crucial for their good generalization performance \cite{valle2018deep}. The general problem of measuring the complexity of deep neural networks has given rise to several complexity metrics \cite{novak2018sensitivity} and studies on how they relate to generalization \cite{jiang2019fantastic}.
	
	Connected to this, recent studies \cite{geiger2019jamming, belkin2019reconciling} on the effect of overparameterization in neural networks led to the rediscovery of the ``double descent'' phenomenon, first observed in the statistical physics literature \cite{opper1995statistical}, which is the observation that when increasing the capacity of a neural network (measured, for example, by the number of parameters) the generalization error shows a sudden peak around the interpolation point (where approximately zero training error is achieved), but then a second decrease towards a low asymptotic value is observed at higher overparameterization. 
	
	In the present work, we study the double descent phenomenon under a notion of function sensitivity based on the \textit{mean dimension} \cite{hahn2022mean, hoyt2021efficient}. The mean dimension yields a measure of the mean interaction order between input variables in a function, and can also be proved to be related to the variance of the function under local perturbations of the input features. 
	While this notion originated in the field statistics \cite{liu2006estimating}, several computational techniques have been proposed for its estimation in the context of neural networks. One of the main obstacles, however, comes from trying to characterize the sensitivity of the function over an input distribution that is strongly structured and not fully known.
	
	In this paper, we propose to focus on the study of the \textit{Boolean mean dimension}~\cite{o2014analysis} (BMD), which involves a simple i.i.d. binary input distribution.
	We show how the BMD can be estimated efficiently, and provide analytical and numerical evidence of the correlation of this metric with several phenomena observed on the data used for training and testing the model.

	\section{Related works}
	\subsection{Overparameterization and Double Descent}
	
	
	Several studies \cite{baity2018comparing, geiger2019jamming, advani2020high} confirmed the robustness of the double descent phenomenology for a large variety of architectures, datasets, and learning paradigms. An analytical study of double descent in the context of the random feature model~\cite{rahimi2007random} was conducted rigorously for the square loss in~\cite{mei2019generalization} and for generic loss by~\cite{gerace2020generalisation} using the replica method~\cite{mezard1987spin}. Double descent has then later found also in the context of one layer model learning a Gaussian mixture dataset~\cite{mignacco2020role}; similarly to the random feature model, the peak in the generalization can be avoided by optimally regularizing the network. In this context in~\cite{Baldassi_2020} it was also shown that choosing the optimal regularization corresponds to maximize a flatness-based measure of the loss minimizer. A range of later studies further explored this phenomenology in related settings~\cite{d2020double, gerace2022probing}. 
	
	Different scenarios have also been shown to give rise to a similar phenomenology, such as the epoch-wise double descent and sample non-monotonicity~\cite{nakkiran2021deep} and the triple descent that can appear with noisy labels and can be regularized by the non-linearity of the activation function \cite{Biroli2020triple}. 
	
	In this work, we connect the usual double descent of the generalization error with the behavior of the mean dimension, which is a complexity metric that can be evaluated without requiring task-specific data.

	\subsection{Mean Dimension and Boolean Mean Dimension}
	
	The \textit{mean dimension} (MD), based on the analysis of variance (ANOVA) expansion \cite{efron1981jackknife, owen2003dimension}, can be intuitively understood as a marker of the complexity of a function due to the presence of interactions between a large set of input variables.
	
	The mean dimension has been used as a tool to analyze and compare for example neural networks \cite{hoyt2021efficient, hahn2022mean} and, with a slightly different definition, also generative models of protein sequences \cite{feinauer2022mean}. The MD has the advantage that it can be calculated for a \textit{black-box function}, without regard to the internal mechanism for calculating the input-output relation. One major drawback, however, is the intense computational cost associated with its direct estimation. This computational limitation has led to the proposal of several approximation strategies ~\cite{hoyt2021efficient, hahn2022mean}. 
	In some special cases, the mean dimension can be explicitly expressed as a function of the coefficients of a Fourier expansion, as seen from the relationship between the Boolean Mean Dimension (BMD) and the \textit{total influence} \cite{o2014analysis} defined in the analysis of Boolean functions (see below), and its generalization \cite{feinauer2022mean} for functions with categorical variables.
	
	\section{Mean Dimension}
	
	In the next paragraphs, we first provide a general mathematical definition of the mean dimension for a square-integrable function with real-valued input distribution. We then specialize to the case of a binary input distribution and define the Boolean Mean Dimension (BMD), which will be the main quantity investigated throughout this paper. 
	Finally, we will discuss how to efficiently estimate the MD and the BMD through a simple Monte Carlo procedure. 
	

	

	\subsection{Mathematical Definition}
	
	To give a proper mathematical definition of the mean dimension, for a real-valued function $f(\boldsymbol{x})$ of $n$ variables $f: \mathbb{R}^n \rightarrow \mathbb{R}$, it is convenient to introduce some notation that will be used in the rest of the paper. We will denote the set of indexes $\{1, \dots \, n\}$ by $[n]$. We define $\boldsymbol{x}_u$ the set of input variables $x_i$, with $i \in u \subseteq [n]$ and by $\boldsymbol{x}_{\backslash u}$ the set of variables for which $i \notin u$. We will also assume that $\boldsymbol{x}$ is drawn from a distribution $p(\boldsymbol{x})$. The basic idea of the mean dimension \cite{hahn2022mean} is to derive a complexity measure for $f$ from an expansion of the type 
	\begin{equation}
		f(\boldsymbol{x}) = \sum_{u \subseteq [n]} f_u(\boldsymbol{x}_u) 
	\end{equation}
	where the ``components'' $f_u(\boldsymbol{x}_u)$ can be computed from the following recursion relation
	\begin{equation}
		\label{eq::ANOVA_recursion}
		f_u(\boldsymbol{x}_u) \equiv \int f(\boldsymbol{x}) \, p(\boldsymbol{x}_{\backslash u} | \boldsymbol{x}_{u}) \, d \boldsymbol{x}_{\backslash u} - \sum_{v \subset u} f_v(\boldsymbol{x}_v)
	\end{equation}
	with the initial condition $f_{\emptyset} = \int f(\boldsymbol{x}) \, p(\boldsymbol{x}) \, d \boldsymbol{x} \equiv \mathbb{E}[f]$. 
	It can be shown that coefficients of the expansion have zero average if $u$ is non empty
	\begin{equation}
		\int f_u(\boldsymbol{x}_u) p_u(\boldsymbol{x}_u) \, d \boldsymbol{x}_u = 0 \, \qquad u \ne \emptyset
	\end{equation}
	where we have denoted by $p_{u}(\boldsymbol{x}_u)$ the marginal probability distribution over the set $u$. Moreover, they satisfy orthogonality relations, namely
	\begin{equation}
		\label{eq::orthogonality}
		\int f_u(\boldsymbol{x}_u) f_{v}(\boldsymbol{x}_v) p_{u \cup v}(\boldsymbol{x}_{u \cup v}) \, d \boldsymbol{x}_{u \cup v} = 0\,, \qquad \text{if} \; u \ne v \,.
	\end{equation}
	Using those relations we can write the variance of the function as a decomposition of $2^n - 1$ terms
	\begin{equation}
		\sigma^2 = \mathbb{E}[f^2] - \mathbb{E}[f]^2 = \sum_{u \subseteq [n] \backslash \emptyset} \sigma_u^2
	\end{equation}
	where 
	\begin{equation}
		\sigma_u^2 \equiv 
		\int f_u^2(\boldsymbol{x}_u) \, p_u(\boldsymbol{x}_u) \, d \boldsymbol{x}_u \,.
	\end{equation}
	The mean dimension $M_f$ is then defined as~\cite{hahn2022mean}
	\begin{equation}
		\label{eq:md}
		M_f = \sum\limits_{u \subseteq [n]} |u| \frac{\sigma^2_u}{\sigma^2},
	\end{equation}
	i.e. a weighted sum over possible interactions, with each subset of inputs contributing based on how much they influence the variance. 
	
	\subsection{Pseudo-Boolean Functions and Fourier coefficients}
	
	We now derive an explicit expression for the mean dimension of $n$-dimensional pseudo-Boolean functions taking values on the real domain, $f: \{-1, 1\}^n \to \mathbb{R}$ under the assumption of input features that are i.i.d. from $\{-1, 1\}$. 
	
	Denoting by $\boldsymbol{s} \in \{-1, 1\}^n$ the $n$-dimensional binary input of $f$, such a function can be uniquely written as a \emph{Fourier expansion}~\cite{o2014analysis} in terms of a finite set of \emph{Fourier coefficients} $\hat f_u$, $u \subseteq [n]$ as
	\begin{equation}
		f(\boldsymbol{s}) = C + \sum_{i} h_i s_i + \sum_{i<j} J_{ij} s_i s_j + \sum_{i<j<k} K_{ijk} s_i s_j s_k + \ldots = \sum_{u \subseteq [n]} \hat{f}_u \, \chi_u(\boldsymbol{s}_u)
	\end{equation}
	where
	\begin{equation}
		\label{eq::boolean_basis}
		\chi_u(\boldsymbol{s}_u) = \prod_{i \in u} s_i
	\end{equation}
	represent the Fourier basis of the decomposition that are orthonormal $\langle \chi_u(\boldsymbol{s}) \chi_{v}(\boldsymbol{s}) \rangle = \delta_{u, v}$ with respect to the uniform distribution over $\{-1, 1\}^n$, where we use the notation
	\begin{equation}
		\label{eq::uniform_boolean_distribution}
		\langle \bullet \rangle \equiv \frac{1}{2^n} \sum_{\boldsymbol{s}\in \{-1, 1\}^n} \bullet \,.
	\end{equation}
	The Fourier coefficients $\hat f_u$ can give information about the moments of the function $f$ with respect to the uniform distribution~\eqref{eq::uniform_boolean_distribution} over $\boldsymbol{s}$; for example the first moment is
	\begin{equation}
		\langle f(\boldsymbol{s}) \rangle = \hat f_{\emptyset}
	\end{equation}
	whereas the variance can be obtained as
	\begin{equation}
		\sigma^2 = 
		\langle f^2(\boldsymbol{s}) \rangle - \langle f(\boldsymbol{s}) \rangle^2 = \sum_{u \subseteq [n] \backslash \emptyset} \hat{f}^2_u \,.
	\end{equation}
	We can quantify the contribution $c_k$ of interaction of order $k$ to the variance of $f(\boldsymbol{s})$ as the ratio
	\begin{equation}
		c_k = \frac{\sum_{u \subseteq [n] \backslash \emptyset : |u| = k} \hat{f}^2_u}{\sigma^2} \,. 
\end{equation}
Notice that $\sum_{k} c_k = 1$, so that $c_k$ can be interpreted as a (discrete) probability measure over interactions. The mean dimension of $f$ can then be written as the mean interaction degree when weighted according to it contribution to the variance, i.e. as a weighted sum of feature influences divided by the total variance of the function, so
\begin{equation}
	\label{eq::mean_dimension}
	M_f \equiv \sum_{k=1}^n k  c_k 
	= \frac{\sum_{u \subseteq [n]} \left| u \right| \hat{f}^2_u}{\sigma^2}
\end{equation}
This expression is equivalent to Eq.~\eqref{eq:md} for pseudo-Boolean functions under the assumptions that all features are i.i.d from $\{-1, 1\}$. 
The expression connects the notion of simplicity in terms of variance contributions to the same notion in terms of explicit expansion coefficients.  Intuitively, a large mean dimension is indicating that the function fluctuates due to a large contribution of high-degree interactions.

\subsection{Estimating the Mean Dimension through Monte Carlo}\label{compute MD}

The expression of the mean dimension in~\eqref{eq:md} involves a sum over all the set of subsets of $n$ variables, and its numerical evaluation through a brute-force approach would be intractable in high dimension. 
However, it can be shown that a more efficient evaluation scheme of equation~\eqref{eq:md}, can be achieved through a Monte Carlo approach~\cite{liu2006estimating}.
First, the MD can be rewritten as a sum over the $n$ input components:
\begin{equation}
	\label{eq::MD_efficient_estimation}
	M_f = \frac{\sum_{i=1}^n \tau_i^2}{\sigma^2}
\end{equation}
where the \emph{influence} of the $i$-th input component $\tau_i$ is defined as:
\begin{equation}\label{MD est}
	\tau_i^2 = \frac{1}{2} \int d \boldsymbol{x} \, d x_i' \, p(\boldsymbol{x}) \,  p(x_i'|\boldsymbol{x}_{\backslash i}) (f(\boldsymbol{x}) - f(\boldsymbol{x}^{\oplus i}))^2 \,. 
\end{equation}
and where we have denoted by $\boldsymbol{x}^{\oplus i}$ a vector $\boldsymbol{x}$ with a resampled $i_{th}$ coordinate. We show an original proof of this identity in Appendix~\ref{app::proof}. 

Note that the definition of the MD for a generic input distribution in Eq.~\eqref{MD est}, entails a resampling procedure that presumes knowledge of the conditional distribution of a pixel given the rest of the pixel values. In the general case, this pixel is to be resampled multiple times from this conditional distribution, to compute the variance of the function under this variation of the input. This conditional distribution, however, is not a known quantity for a real dataset. For this reason, for example, some authors have proposed an “exchange” procedure, where one randomply samples a different pixel value observed in the same dataset \cite{hahn2022mean}, however this approximation neglects the within sample correlations. 


Expression \eqref{MD est} can be specialized to the case of binary i.i.d. inputs, where one can identify the influence functions $\tau_i^2$ with the discrete derivatives:
\begin{equation} \label{eq::taudiscrete}
	\tau_i^2 = \left\langle \left( \mathcal{D}_i f(\boldsymbol{s}) \right)^2 \right\rangle
\end{equation}
where $\mathcal{D}_i f(\boldsymbol{s})$ denotes the $i_{\text{th}}$ (discrete) derivative of $f(\boldsymbol{s})$, i.e.
\begin{equation}
	\mathcal{D}_i f(\boldsymbol{s}) \equiv \frac{f(s_1, \dots, s_i = 1, \dots, s_n) - f(s_1, \dots, s_i=-1, \dots, s_n)}{2}
\end{equation}
and measures the average sensitivity of the function to a flip of the $i_{\text{th}}$ variable. 
The sum of the influences $\sum_i \left\langle \left( \mathcal{D}_i f(\boldsymbol{s}) \right)^2 \right\rangle$ 
is known in the field of the analysis of pseudo-Boolean functions as \emph{total influence} of $f$~\cite{o2014analysis}. 
In terms of the Fourier expansion, we have
\begin{equation}
	\mathcal{D}_i f(\boldsymbol{s}) = \sum_{u \subseteq [n]: i \in u} \hat f_u \, \chi_{u \backslash i}(\boldsymbol{s}_{u \backslash i})
\end{equation}
Therefore computing the mean dimension for pseudo-Boolean functions boils down to querying the function $f$ on uniformly sampled binary sequences of length $n-1$. 

\subsection{Boolean Mean Dimension}

In the general case, the underlying input distribution of the training dataset is not known and estimating the MD on this distribution becomes unfeasible. In the present work, we propose employing the estimation procedure presented in the last section, based on binary sequences, as an easily computable proxy of the sensitivity of the neural network function.
In order to distinguish this proxy from the mean dimension over the dataset distribution, we call the resulting quantity the Boolean Mean Dimension (BMD). We show in the results below that the BMD can in some cases be computed analytically, and that it is qualitatively related to the generalization phenomenology in neural networks.

\section{Analytical results}

We now derive an analytic expression for the mean dimension in the special case of the random feature model~\cite{rahimi2007random,goldt2019modelling,loureiro2021capturing,baldassi2022learning}, focusing on the same high dimensional regime where the double descent phenomenon can be detected. In the next sections, we will define the model, the learning task and the high dimensional limit precisely, and we will sketch the analytical derivation of the expression for the Boolean Mean Dimension.

\subsection{Model definition and learning task}

The random feature model (RFM) is a two-layer neural network with random and fixed first-layer weights (also called features) and trainable second-layer weights. Given a $D$-dimensional input, $\boldsymbol{x}\in\mathbb{R}^D$, and denoting by $F\in\mathbb{R}^{D\times N}$ the $D\times N$ frozen feature matrix, the pre-activation of the RFM is given by:
\begin{equation}
	\hat y(\boldsymbol{w}; \boldsymbol{x}) = \frac{1}{\sqrt{N}} \sum_{i=1}^N w_i \, \sigma\left( \frac{1}{\sqrt{D}} \sum_{k=1}^D F_{ki} \, x_k \right) 
\end{equation}
where $\boldsymbol{w}$ is an $N$-dimensional weight vector and $\sigma$ is a (usually non-linear) function. The parameter $N$ indicates the number of features in the RFM and can be varied to change the degree of over-parametrization of the model. 
As in~\cite{baldassi2022learning}, we will hereafter focus on the case of i.i.d. standard normal distributed feature components $F_{ki}\sim\mathcal{N}(0,1)$, although the formalism allows for a simple extension to a generic fixed feature map, under a simple weak correlation requirement (see~\cite{gerace2020generalisation, loureiro2021capturing} for additional details).

We consider a classification task defined by a training dataset of size $P$, denoted as $\mathcal{D} = \left\{\boldsymbol{x}^\mu, y^\mu\right\}_{\mu = 1}^P$. The inputs are assumed to be i.i.d. with first and second moments fixed respectively to $\mathbb{E} x_i = 0$ and $\mathbb{E} x_i^2 = 1$. Note that, for example, both binary input components $x_i \in \{-1,1\}$ and Gaussian components $x_i \sim \mathcal{N}(0,1)$ satisfy the above assumption. The binary labels $y^\mu\in\{-1,1\}$ are assumed to be produced by a ``teacher'' linear model $\boldsymbol{w}^T\in\mathbb{R}^D$, with normalized weights on the $D$-sphere $\lVert \boldsymbol{w}^T \rVert^2_2 = D$, according to:
\begin{equation}
	y^\mu = \text{sign}\left(\frac{1}{\sqrt{D}} \sum_{k=1}^D w_k^T x_k^\mu \right)\,, \qquad \mu \in [P] \,.
\end{equation}
The learning task is then framed as an optimization problem with generic loss function $\ell$ and ridge regularization
\begin{equation}
	\label{eq::learning_task}
	\boldsymbol{w}_\star = \argmin_{\boldsymbol{w} \in \mathbb{R}^N}\left[ \sum_{\mu = 1}^P \ell (y^\mu , \, \hat y^\mu (\boldsymbol{w}; \boldsymbol{x}^\mu)) + \frac{\lambda}{2} \lVert \boldsymbol{w} \rVert^2_2 \right] \,,
\end{equation}
where $\lambda$ is a positive external parameter controlling the regularization strength. In the following we will consider the two most common convex loss functions, namely the mean squared error (MSE) and the cross-entropy (CE) losses, defined as
\begin{subequations}
	\label{eq::loss_functions}
	\begin{align}
		\ell_{mse}(y, \hat y) &= \frac{1}{2} \left(y^\mu - \hat y^\mu \right)^2 \\
		\ell_{ce}(y, \hat y) &= \log \left( 1 + e^{- y \hat y } \right)\,.
	\end{align}
\end{subequations}
We analyze the learning problem in the high-dimensional limit where the number of features, input components and training-set size diverge $N\,, \, D\,, \, P\to \infty$ at constant rates $\alpha \equiv P/N = \mathcal{O}(1)$ and $\alpha_D \equiv D/N = \mathcal{O}(1)$. In this limit, strong concentration properties allow for a deterministic characterization of the above-defined learning problem in terms of a finite set of scalar quantities called order parameters. In the next sections, and in detail in the appendices, we will sketch the derivation of this reduced description. 

\subsection{Rephrasing the problem in terms of the Boltzmann measure}

The learning task in~\eqref{eq::learning_task} can be characterized within a statistical physics framework. One can introduce a probability measure over the weights $\boldsymbol{w}$ in terms of the Boltzmann distribution
\begin{equation}
	\label{eq::boltzmann_measure_w}
	\boldsymbol{w} \sim p_\beta(\boldsymbol{w}; \mathcal{D}) = \frac{e^{- \beta \sum_{\mu = 1}^P \ell (y^\mu , \, \hat y^\mu (\boldsymbol{w}; \boldsymbol{x}^\mu)) - \frac{\beta \lambda}{2} \sum_{i=1}^N w_i^2} }{Z_\beta}
\end{equation}
where $\beta$ is the inverse temperature, the loss function in~\eqref{eq::learning_task} plays the role of an energy, and the partition function $Z_\beta$ is a normalization factor that reads
\begin{equation}
	Z_\beta = \int d \boldsymbol{w} \, e^{- \beta \sum_{\mu = 1}^P \ell(y^\mu , \, \hat y^\mu (\boldsymbol{w}; \boldsymbol{x}^\mu)) - \frac{\beta \lambda}{2} \sum_{i=1}^N w_i^2} \,.
\end{equation}
The distribution $p_\beta(\boldsymbol{w}; \mathcal{D})$ can be interpreted in a Bayesian setting as the posterior distribution over the weights $\boldsymbol{w}$ given a dataset $\mathcal{D}$, and~\eqref{eq::boltzmann_measure_w} corresponds to Bayes theorem, where the term $e^{-\beta \sum_\mu \ell (y^\mu , \, \hat y^\mu (\boldsymbol{w}; \boldsymbol{x}^\mu))}$, corresponds to the likelihood and $e^{-\frac{\beta \lambda}{2} \lVert \boldsymbol{w} \rVert_2^2}$ is the prior distribution over the weights. 

In the zero-temperature limit, when $\beta \to \infty$, the probability measure $p_{\beta}(\boldsymbol{w}; \mathcal{D})$ concentrates on the solutions to the optimization problem in~\eqref{eq::learning_task}. To characterize the typical (i.e. the most probable) properties of these solutions, one needs to perform an average over the possible realizations of the training set $\mathcal{D}$ and of the features $F$, computing the free-energy of the system
\begin{equation}
	\label{eq::free_energy}
	f = -\lim_{\beta \to \infty} \lim_{N \to \infty} \frac{1}{\beta N}\mathbb{E}_{\mathcal{D}, F} \ln Z_\beta\ \,.
\end{equation}
The computation of this ``quenched'' average can be achieved via the replica method~\cite{mezard1987spin} from spin-glass theory, which reduces the characterization of the solutions of~\eqref{eq::learning_task} to the determination of a finite set of scalar quantities called order parameters~\cite{engel2001statistical, malatesta2023high}.

In appendix~\ref{app::free_entropy}, we sketch the replica calculation for the free energy, first presented in \cite{gerace2020generalisation}, in the simplifying case of an odd non-linear activation $\sigma$. 

\subsection{Analytical determination of the BMD in the RFM}

We now derive an analytic expression for the Boolean Mean Dimension (BMD) which can be efficiently evaluated for a trained RFM. The definition~\eqref{eq::MD_efficient_estimation} reads
\begin{equation}
	M_f(\boldsymbol{w}) \equiv \frac{ \frac{1}{2}\sum_{k=1}^D \left \langle (\hat{y}(\boldsymbol{w}; \boldsymbol{x}) - \hat{y}(\boldsymbol{w}; \boldsymbol{x}^{\oplus k}))^2 \right \rangle}{\left \langle \hat{y}^2(\boldsymbol{w}; \boldsymbol{x}) \right \rangle - \left \langle \hat{y}(\boldsymbol{w}; \boldsymbol{x}) \right \rangle^2} \,.
\end{equation}
where $\left \langle \bullet \right \rangle$ and $x^{\oplus k}$, defined in ~\eqref{eq::uniform_boolean_distribution} and ~\eqref{MD est}, entail an expectation over i.i.d. uniform binary inputs. In appendix~\ref{app::BMD}, we perform the annealed averages appearing in the numerator and the denominator separately, obtaining the expression:   
\begin{equation} 
	M_f(\boldsymbol{w}) = \frac{\frac{1}{N} \sum_{ij} \bar{\Psi}_{ij} w_i w_j}{\frac{1}{N} \sum_{ij} \Psi_{ij} w_i w_j} \label{eq::MD_high-dim}
\end{equation}
where we defined
\begin{subequations}
	\begin{align}
		\Omega_{ij} &\equiv \frac{1}{D} \sum_{k=1}^D F_{ki} F_{kj} \,. \\
		\bar{\Psi}_{ij} &\equiv \bar{\kappa}_\star^2 \, \Omega_{ii} \mathbb{I}_{ij} + \bar{\kappa}_0^2 \, \Omega_{ij} + \bar{\kappa}_1^2 \, \Omega^2_{ij} \,,  \\
		\Psi_{ij} &\equiv \kappa_\star^2 \, \mathbb{I}_{ij} + \kappa_1^2 \, \Omega_{ij} \,. 
	\end{align}
\end{subequations}
and the coefficients $\kappa$ are defined as expectations of derivatives of the activation function over a standard Gaussian measure $Dz=\frac{e^{-z^2/2}}{\sqrt{2\pi}}dz$:
\begin{subequations}
	\begin{align}
		\kappa_0 &= \int D z \, \sigma(z)\,, & \kappa_1&=\int Dz\, \sigma^\prime(z)\,, & \kappa_2&=\int Dz \, \sigma^2(z) \,,\\
		\bar{\kappa}_0 &= \kappa_1\,, & \bar{\kappa}_1&=\int Dz \, \sigma^{\prime\prime}(z)\,, &  \bar{\kappa}_2 &= \int Dz \, \left(\sigma^{\prime}(z) \right)^2\,,   \\
		\kappa_\star^2 &= \kappa_2 - \kappa_1^2 - \kappa_0^2 \,, & \bar{\kappa}_\star^2 &= \bar{\kappa}_2 - \bar{\kappa}_1^2 - \bar{\kappa}_0^2 \,. &
	\end{align}
\end{subequations}
As we show in the appendix, the above expression~\eqref{eq::MD_high-dim} is universal: evaluating the MD with respect to a different i.i.d. input distributions with matching first and second moments would give exactly the same result. 

Moreover, note that the evaluation of expression~\eqref{eq::MD_high-dim} no longer involves a Monte-Carlo over the input distribution, with a major gain in computational cost. In appendix~\ref{app::BMD}, we show the agreement of this compact formula with the computationally more expensive Monte Carlo estimation of the BMD.

\begin{figure}[H]
	\captionsetup{justification=raggedright, singlelinecheck=false}
	\includegraphics[width=0.5\textwidth]{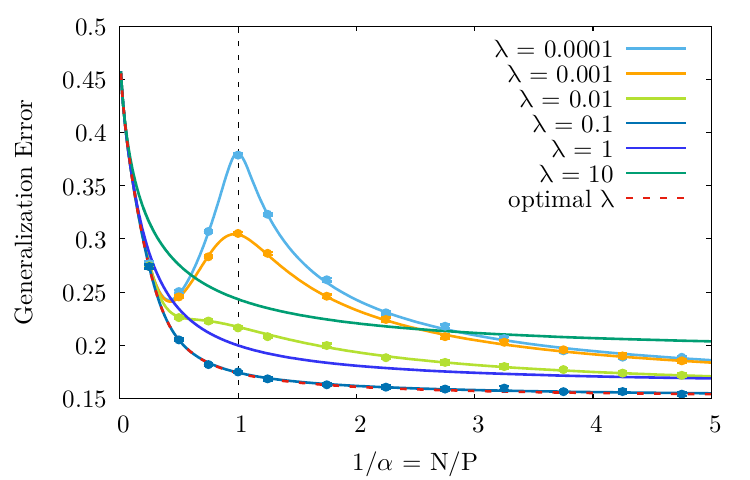}
	\includegraphics[width=0.5\textwidth]{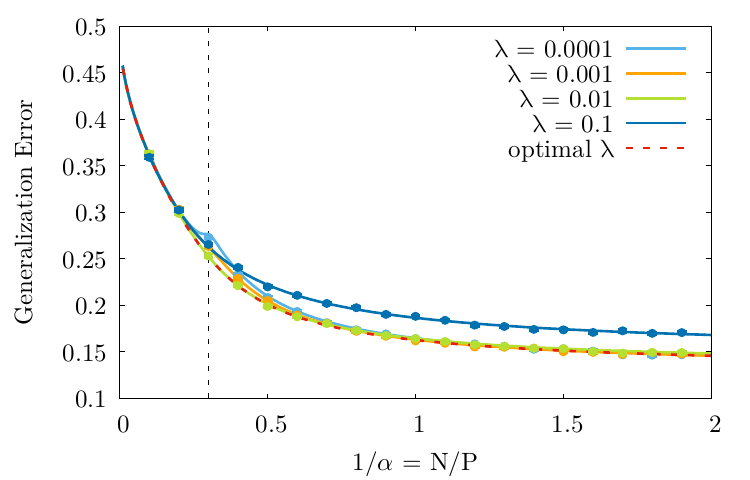}
	\includegraphics[width=0.5\textwidth]{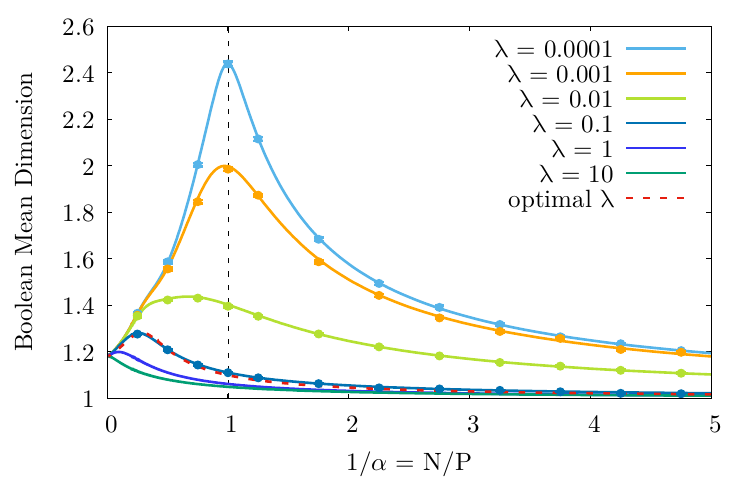}
	\includegraphics[width=0.5\textwidth]{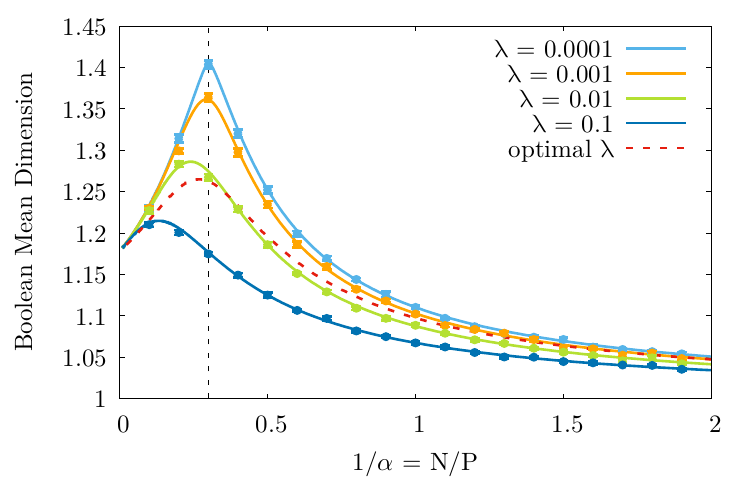}
	\caption{Generalization error (top panels) and BMD (bottom panels) as a function of the overparameterization degree $1/\alpha = N/P$, for fixed $\alpha_T = P/D = 3$ and with $\sigma = \tanh$. The left and right panels represents respectively the case of the MSE and of the CE loss. Several value of the regularization $\lambda$ are displayed, together with the optimal one (which was found by minimizing the generalization error for each value of $\alpha$, see red dashed line). As it can be seen in both plots, for small regularization $\lambda$, the location of the peak in the generalization error exactly coincides with the one in the BMD (vertical dashed lines). As one increases the regularization the peak in the both the generalization and the BMD is milded. }
	\label{fig:MD_analytic}
\end{figure}
The mean dimension therefore explicitly depends on the model parameters $\boldsymbol{w}$. The evaluation of the typical BMD of a trained RFM can thus be computed by taking an expectation over the zero-temperature Boltzmann measure for the weights derived in the replica computation, $M_f = \mathbb{E}_{\mathcal{D}, F} \left\langle M_f(\boldsymbol{w}) \right\rangle_{\boldsymbol{w}}$. 
The notation $\left \langle \bullet \right \rangle_{\boldsymbol{w}} \equiv \int d\boldsymbol{w} \, \bullet p_{\infty}(\boldsymbol{w}; \mathcal{D})$ is thus used to indicate an average over the posterior distribution in equation~\eqref{eq::boltzmann_measure_w}, in the large $\beta$ limit. 

In the case of the replica computation for an odd activation function, that we reported in appendix~\ref{app::free_entropy}, one can simplify further expression \eqref{eq::MD_high-dim} by recognizing that $\bar{\kappa}_1=0$ and that $\Omega_{ii}=1$ when the feature components have second moment equal to $1$. In this case, the numerator and the denominator can be directly expressed in terms of the order parameters of the model:
\begin{equation}
	M_f = 1 + \left(\bar{\kappa}_2 - \kappa_2\right) \frac{q_d }{Q_d}
\end{equation}
where 
\begin{subequations}
	\begin{align}
		q_d &\equiv \mathbb{E}_{\mathcal{D}, F} \left\langle \frac{1}{N} \sum_{i=1}^{N} w_i^2 \right\rangle_{\boldsymbol{w}}\,, \\
		p_d &\equiv \mathbb{E}_{\mathcal{D}, F} \left\langle \frac{1}{N} \sum_{i,j = 1}^{N} \Omega_{ij} w_i w_j \right\rangle_{\boldsymbol{w}}\,, \\
		Q_d &\equiv \kappa_\star^2 q_d + \kappa_1^2 p_d \,.
	\end{align}
\end{subequations}
The order parameters $q_d$, $p_d$ can be computed by solving saddle point equations as shown in Appendix~\ref{app::free_entropy}. 

Notice that in the case of a linear activation function the BMD is always 1 since a flip in the inputs will induce always the same response. 

In Fig.~\ref{fig:MD_analytic} we show the plot of the generalization error and the corresponding BMD of the RFM at a fixed $\alpha_T$, as a function of $1/\alpha$ for the MSE (left panels) and CE loss (right panels). As shown in~\cite{gerace2020generalisation}, for small regularization $\lambda$ the generalization error develops a peak approximately where the model starts to fit all training data. In the case of the MSE loss, this threshold is often called interpolation threshold and it is located at $N=P$. When using the CE loss, this happens when the projected data become linearly separable and the exact location of the threshold strongly depends on the input statistics and features. Exactly in the correspondence of the generalization error peak the BMD displays its own peak, meaning that the function implemented by the network is more sensitive to perturbation of the inputs. 	

An interesting insight can be deduced from the behavior of the BMD at the optimal value of regularization for the RFM (dashed red curves in Fig.~\ref{fig:MD_analytic}). While the generalization error becomes monotonic as the over-parametrization is increased, the BMD still reaches a peak at first and then descends to $1$ only in the kernel limit $N/P\to\infty$. This might be surprising since the ground-truth linear model, the teacher, has BMD equal to $1$ and one would expect the best generalizing RFM to achieve the best possible approximation of this function and therefore to match its BMD. However, blind minimization of the BMD is not compatible with good generalization, as seen from the performance of the RFM with very large regularization $\lambda$. The explanation of this comes from the architectural mismatch between the linear teacher and the RFM: according to the GET the RFM learning problem is equivalent to a linear problem with an additional noise with an intensity regulated by the degree of non-linearity of the activation function~\cite{Biroli2020triple}. This noise initially forces the under-parameterized RFM to overstretch its parameters to fit the data, causing an increased sensitivity to input perturbations. As the over-parameterization is increased, the RFM becomes equivalent to an optimally regularized linear model~\cite{gerace2020generalisation} and the BMD slowly drops to 1 in this limit. 

Note that in the large dataset limit, when $\alpha,\alpha_T\to\infty$ with $\alpha_D=\mathcal{O}(1)$, a secondary peak for the BMD of the RFM emerges around $\alpha_D=1$, i.e. when the number of parameters of the RFM is the same as the number of input features. This peak is caused by the insurgence of singular values in the spectrum of the covariance matrix $\Omega$ and is more accentuated at lower values of the regularization. Since modern deep networks operate in a completely different regime from the large dataset limit specified above, we expect this secondary peak not to be visible in realistic settings. For example, in the above plots in the low regularization regime, this peak is overshadowed by the main BMD peak. We analyze this phenomenology in detail in appendix~\ref{app::BMD_two_peaks}.

\section{Numerical results}

In the following subsections, we explore numerically the robustness of the BMD phenomenology analyzed in the RFM, considering different types of data distribution, model architecture and learning task. 

Furthermore, we show that adversarially initialized models also display higher BMD, and that the increased sensitivity associated with a large BMD can hinder the robustness of the model against random perturbations of the training inputs.

Finally, we show that the location of the BMD peak is robust to the choice of input statistics used for its measurement, even in non-i.i.d. settings.

\subsection{Experimental setup}\label{Experimental setup}

In the following subsections, each panel displays the performance of a large number of different model architectures with varying degree of over-parameterization, trained on different datasets. Except where specified otherwise, all model are initialized with the common \textit{Xavier} method~\cite{glorot2010xavier} and use the Adam optimizer~\cite{kingma2014adam}, with batch size $128$ and learning rate $10^{-4}$. No specific early stopping criterion is implemented. As in other works analysing the double descent, we experiment with different levels of uniformly random label noise during training (which is introduced by corrupting a random fraction of labels), which tends to make the double descent peak more pronounced~\cite{nakkiran2021deep}. We discuss the effect of label noise below.

\subsubsection{Model architectures}

We consider different types of model architectures: 
\begin{itemize}
	\item Random feature model (RFM), described above, where the number of hidden neurons in the first (fixed) layer controls the degree of over-parameterization.
	\item Two-layer fully-connected network (MLP) with tanh activation, where the number of hidden neurons in the first layer controls the degree of over-parameterization.  
	\item ResNet-18: a family of minimal ResNet \cite{he2016deep} architectures based on the implementation of \cite{nakkiran2021deep}. The structure is finalized with fully connected and softmax layers. As in \cite{nakkiran2021deep},  we control the over-parameterization of the model by changing the number of channels in the convolutional layers. Namely, the 4 ResNet blocks contain convolutional layers of widths [$k$, $2k$, $4k$, $8k$], with $k$ varying from $1$ to $20$.
\end{itemize}  
Both RFM and two-layer fully connected networks in our experiments use hyperbolic tangent activation functions and have weights initialized from a Gaussian distribution and bias terms initialized with zeros. The loss function optimized during training is the cross-entropy loss with $L_2$ regularization (the intensity of the regularization is set to zero if not specified otherwise).

\subsubsection{Data preprocessing}

In the following experiments, we use continuous inputs during the training of the models, normalizing the input features to lie within the $\left[-1, 1 \right]$ interval. While such normalizations are common in preprocessing pipelines, here this procedure has also the benefit of matching the range of variability of the training inputs with that of the randomly i.i.d. sampled binary sequences used to estimate the BMD. We explore the effect of different normalization ranges in Appendix Sec.~\ref{sec:data_range}.

\subsection{MD and generalization peaks as a function of overparametrization}

\begin{figure}[H]
	\captionsetup{justification=raggedright, singlelinecheck=false}
	\includegraphics[scale=0.5]{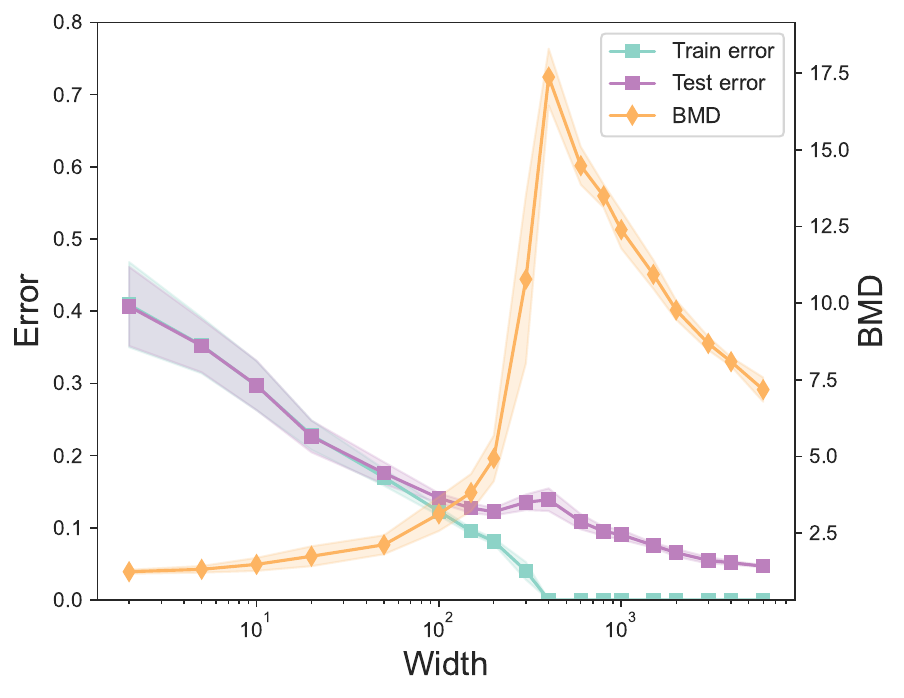}
	\includegraphics[scale=0.5]{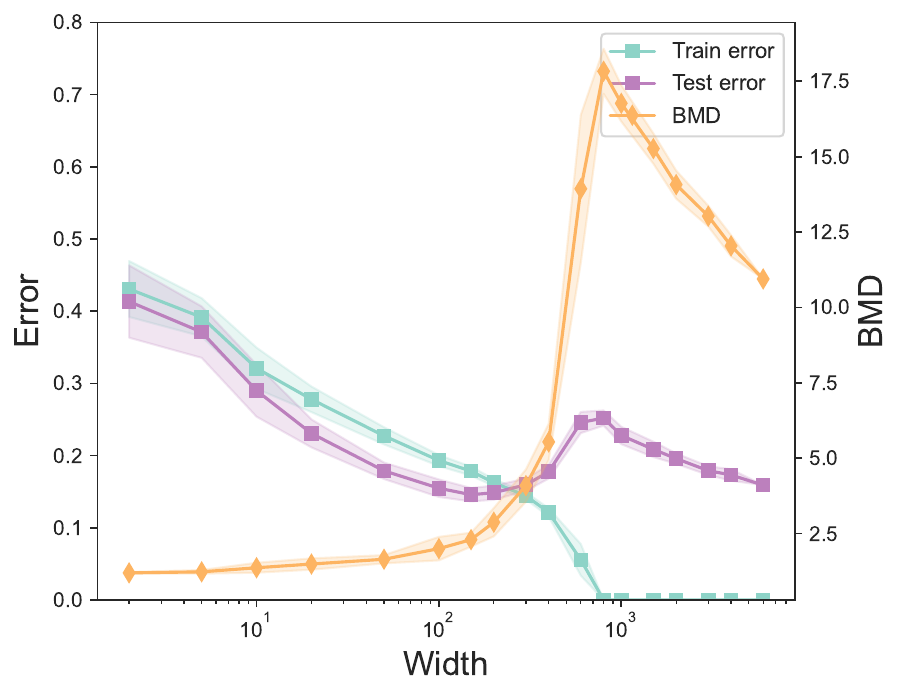}
	\caption{Train error (turquoise), test error (violet) and BMD (orange) curves of the random feature model trained on the MNIST dataset with binary labels on 5K train samples with 0\% label noise (left) and 10\% label noise (right), tested on 5K samples. The resulting plots represent an average (and standard deviations) obtained repeating 20 different times the experiment.}
	\label{fig:noise1}
\end{figure}

\begin{figure}[H]
	\captionsetup{justification=raggedright, singlelinecheck=false}
	\includegraphics[scale=0.5]{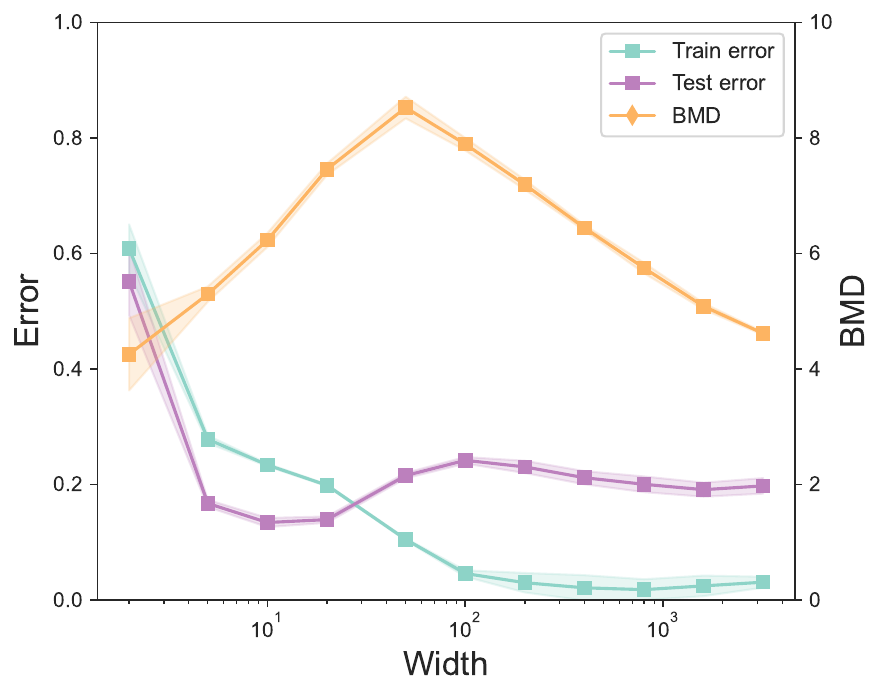}
	\includegraphics[scale=0.5]
	{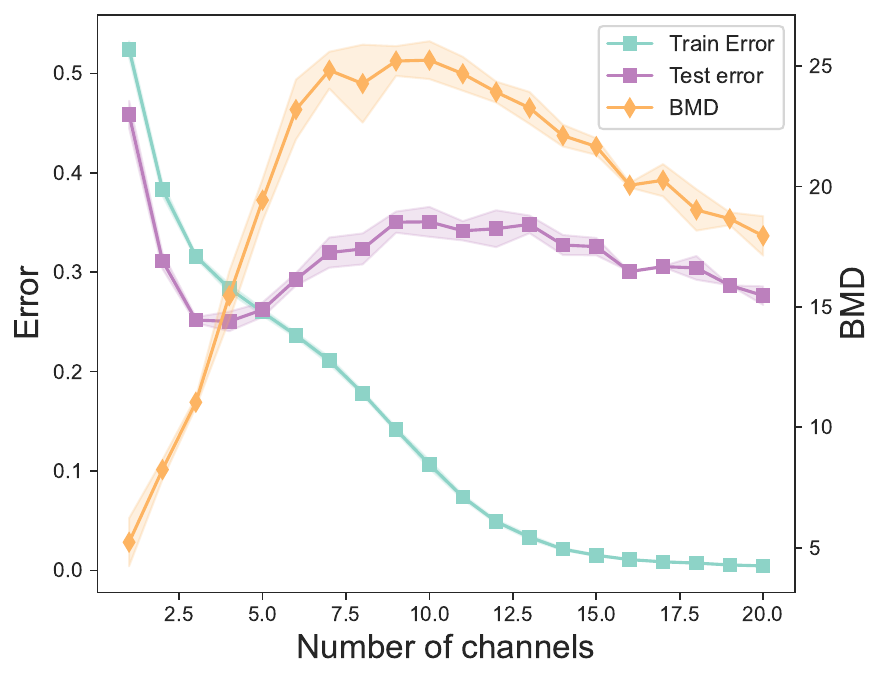}
	\caption{(Left) Train error (turquoise), test error (violet) and BMD (orange) of the two-layer fully-connected network trained on the MNIST dataset with 10 labels on 20K train samples with 20\% label noise, tested on the 5K samples. The resulting plot represents an average (and standard deviations) obtained repeating 20 different times the experiment. (Right) Train error, test error and BMD  of ResNet-18 trained on the CIFAR-10 with 15\% label noise in the train set. The resulting plot represents an average (and standard deviations) obtained repeating 5 different times the experiment.}
	\label{fig:noise2}
\end{figure}

In Fig.~\ref{fig:noise1} we show train and test error, and the BMD for an RFM trained with and without label noise on binary MNIST (even vs odd digits) as a function of the hidden layer width.  
In Fig. \ref{fig:noise2}, we instead consider a two-layer MLP trained on 10-digits MNIST (varying width) and a ResNet-18 trained on CIFAR10 (varying number of channels), both with label noise. In the multi-label case, we are defining the BMD of the network as the average of the BMDs over the classes, where the output of the network is a vector of predicted log-probabilities for each class (i.e. there is a log-softmax activation in the last layer).  

\paragraph*{Position of the BMD peak} The BMD displays a peak around the point where the number of parameters of the model allows it to reach zero training error, in close correspondence with the generalization error peak. We find this phenomenology to be robust with respect to the model class, the dataset, and the over-parameterization procedure. Notice however, that standard optimizers based on SGD are able to implicitly regularize the trained models and can strongly reduce the peaking behavior, as already observed in the context of double descent. In the presented figures we introduced label-noise, which ensures the presence of over-fitting and is thus able to restore both peaks. 

An important observation is that, in order to see this phenomenology, it is not necessary to account for the training input distribution for the evaluation of the MD, which would not be possible in the case of real data. 
In fact, in the over-fitting regime, it is possible to detect an increased sensitivity of the neural network function for multiple input distributions, including the i.i.d. binary inputs entailed in the BMD evaluation. This is explored further in sub-section \ref{sec::input_distribution}.

\paragraph*{Asymptotic behavior of the BMD} When the degree of parametrization of the model is further increased, the BMD decreases and settles on an asymptotic value. The decrease of the BMD in the number of parameters is faster with lower label noise, see Fig.\,\ref{fig:noise1} (left panel vs. right panel). The asymptotic value, reached in the limit of an infinite number of parameters, is task- and model-dependent. For example, in Fig.~\ref{fig:noise1}, the functions learned by the RFMs no longer approximate a linear model (BMD equal to $1$), and are instead bound to higher values of the BMD. 

\paragraph*{Visibility of the BMD peak and Label Noise} The double-descent generalization peak can be a very subtle phenomenon when the learning task is too coherent and the noise level in the data is too weak. With this type of data, the phenomenon can be made more evident \cite{nakkiran2021deep} by adding label noise to the training data. This strategy naturally reduces the signal-to-noise ratio and increases the over-fitting potential during training. The BMD peak, however, seems to be easily identifiable even with zero label noise, (see left panel of Fig.\,\ref{fig:noise1}) where the generalization peak is less pronounced. Note that the BMD does not require any data (neither training nor test) in order to be estimated, so it can be used as a black-box test for assessing the proximity to the separability threshold and therefore as a signal of over-fitting.

\begin{figure}[H]
	\includegraphics[scale=0.38]{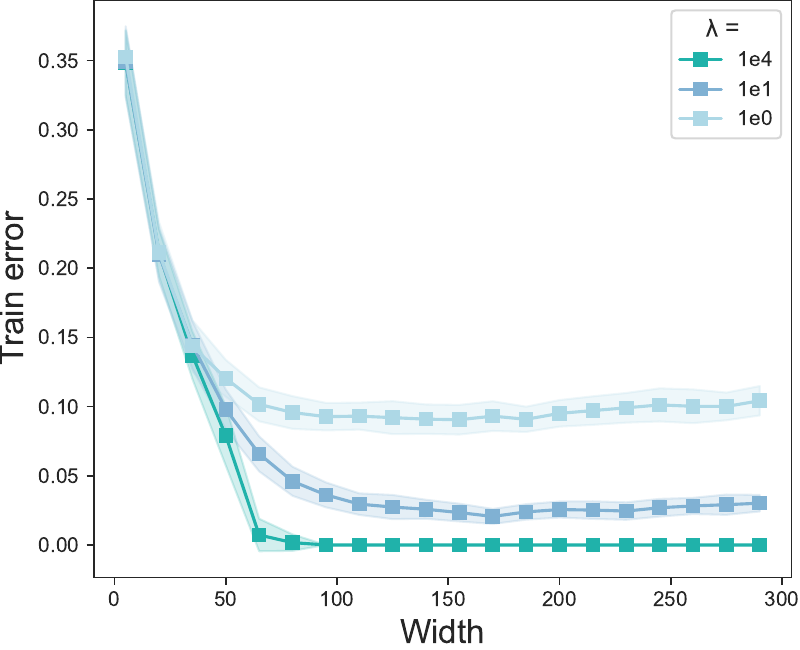}
	\includegraphics[scale=0.38]{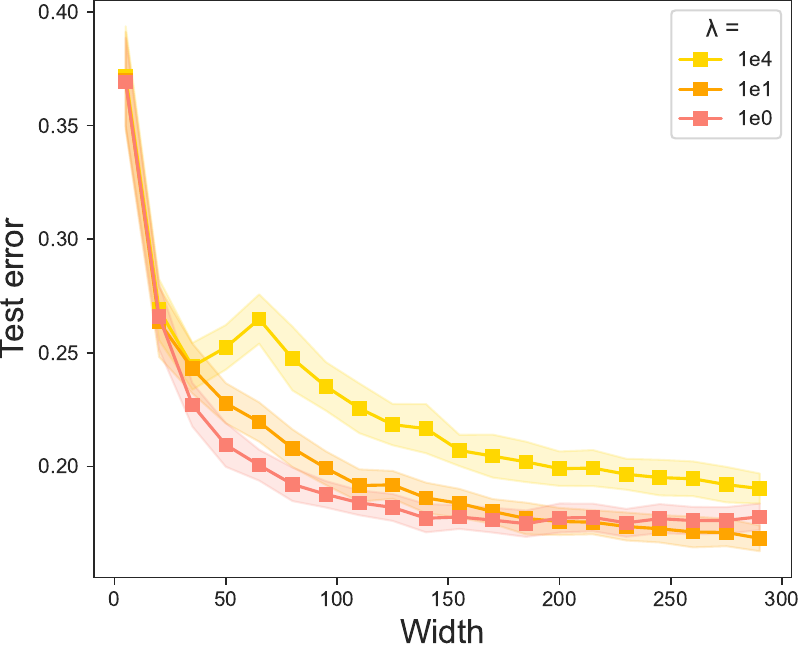}
	\includegraphics[scale=0.38]{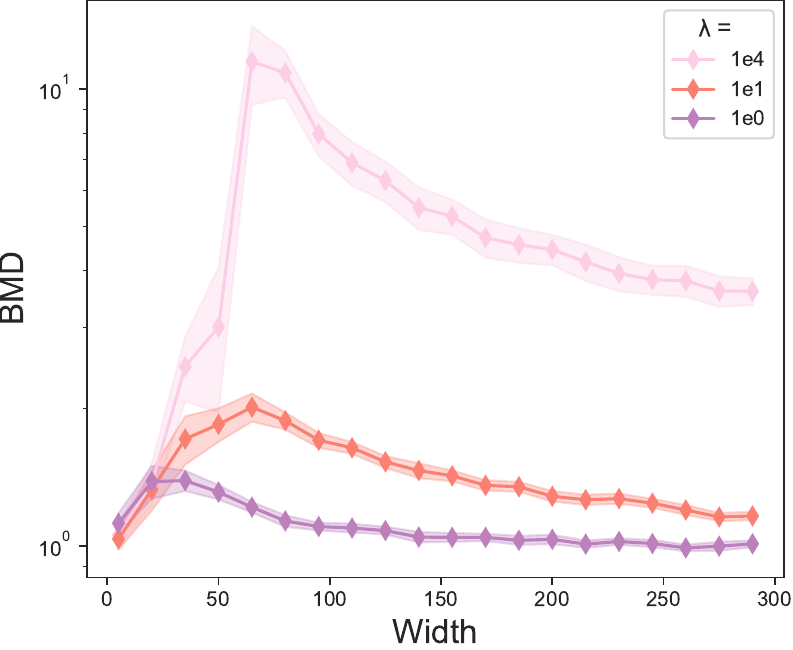}
	\caption{Impact of training with L2 regularization of the random feature model using the MNIST dataset with 10 labels, 200 train samples with no label noise and evaluating on 5K test samples. The loss used is cross-entropy. In all plots, the curves are colored by the strength of the regularization weight $\lambda$. (Left) Regularization effect on the train error.
		(Center) Regularization effect on the test error.
		(Right) Regularization effect on the BMD. The generalization error smoothly decreases with the degree of over-parameterization. Similarly, the BMD peak can be dampened by adding stronger regularization.}
	\label{fig:regularization}
\end{figure}

\paragraph*{Impact of regularization} 
It has been shown that regularizing the model weakens the double-descent peak and that, at the optimal value of the regularization intensity, the generalization error smoothly decreases with the degree of over-parameterization. Similarly, the BMD peak can be dampened by adding stronger regularization, as shown in Fig.\,\ref{fig:regularization}. 


\subsubsection{BMD and Training Set Size}

\begin{center}
	\begin{figure}[H]
		\includegraphics[scale=.55]{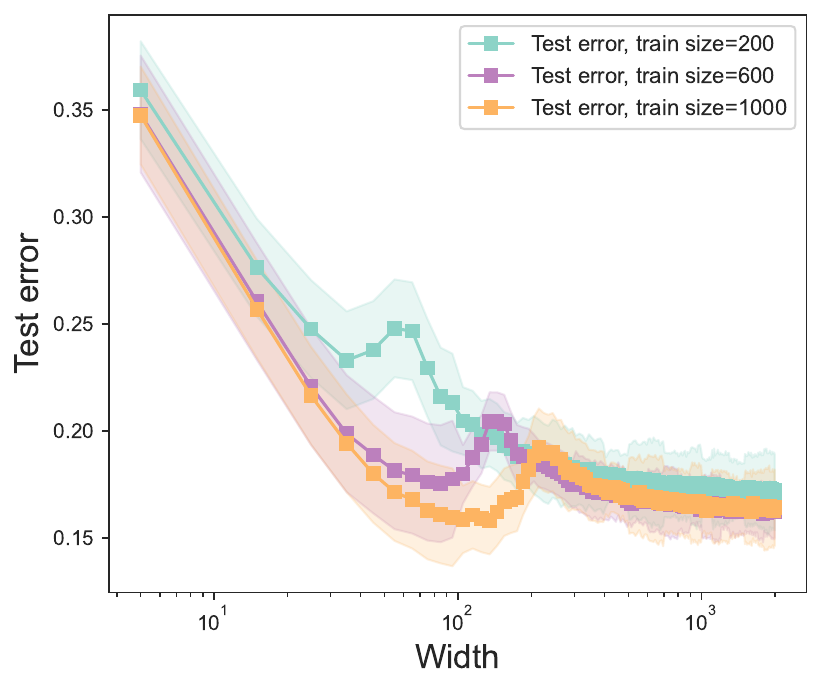}
		\includegraphics[scale=0.55]{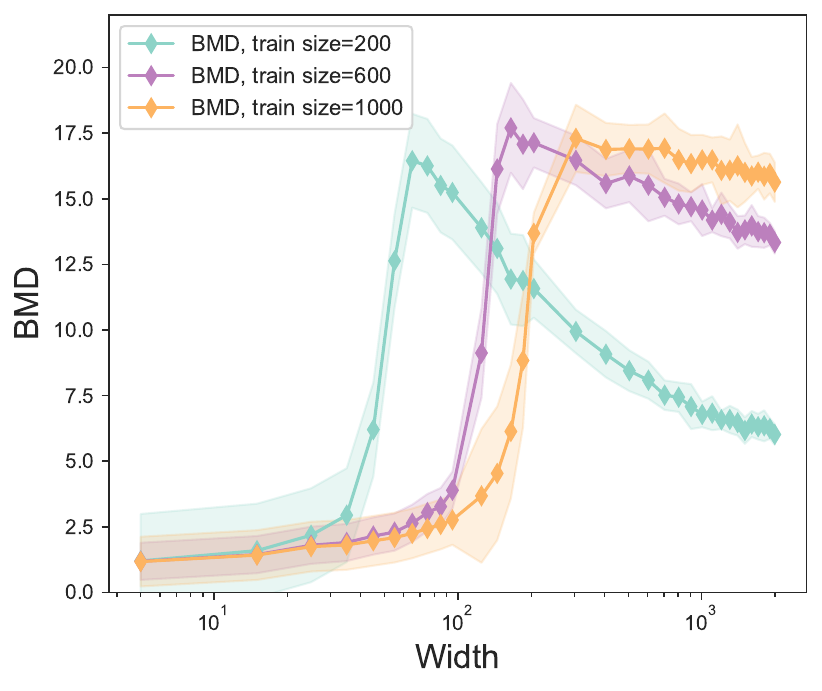}
		\caption{Effect of changing the training set size on the test error (left panel) and the BMD (right panel) of the random feature model using the MNIST dataset with 10 labels, with no label noise and evaluating on 200 test samples. The resulting plot represents an average (and standard deviations) obtained repeating 15 different times the experiment.}
		\label{fig:diff_train_size}
	\end{figure}
\end{center}

In this section, we investigate the effect of varying the number of training samples for a fixed model capacity and training procedure. By increasing the number of training samples, starting from a low number, the same model can switch from being over- to under-parameterized. Therefore increasing the number of training samples has two effects on the test error curve: on the one hand, increasing the number of training samples decreases the test error, shifting the test error curve mostly downwards. On the other hand, increasing the number of training samples increases the capacity at which the double descent peak occurs since a higher capacity is needed until the training set is effectively memorized. This shifts the test error curve (and the BMD curve) to the right. This effect can be seen in Figure \ref{fig:diff_train_size}.

\subsection{BMD and Adversarial Initialization} \label{sec:adv}

In this section, we analyze the BMD of two-layer fully connected networks under adversarial initialization ~\cite{liu2020bad} on the MNIST dataset. This initialization scheme can be used to artificially hinder the generalization performance of the model, forcing it to converge on a bad minimum of the loss. We here aim to show that the initialization has also an effect on the BMD of the model, increasing the sensitivity of the network.

\begin{center}
	\begin{figure}[H]
		\includegraphics[scale=.55]{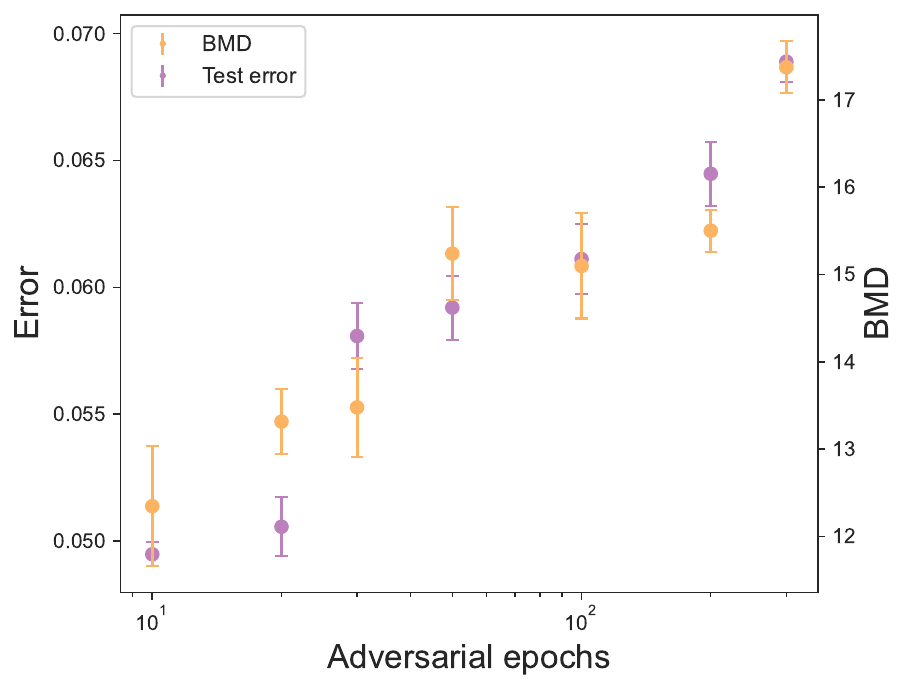}
		\includegraphics[scale=0.55]{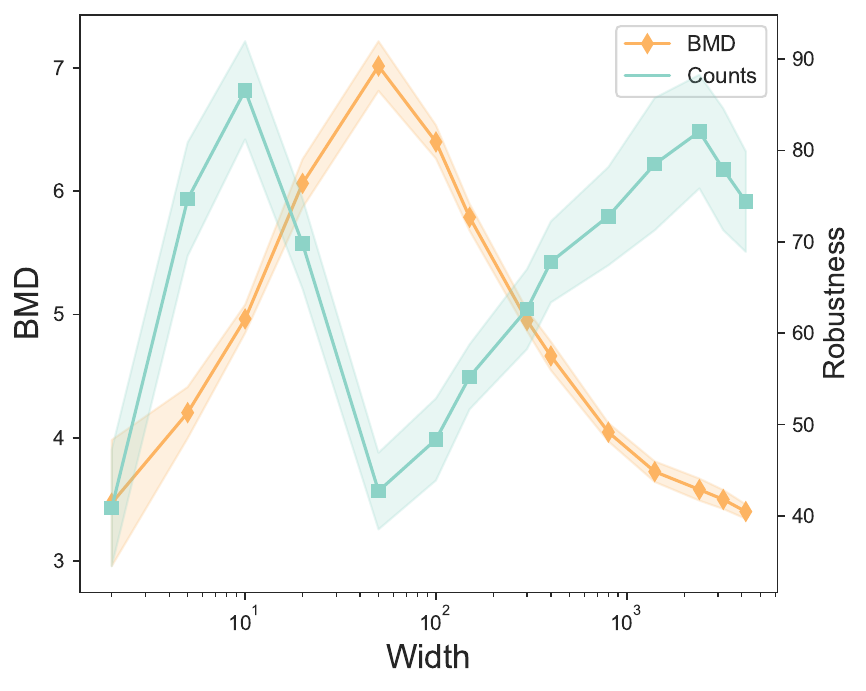}
		\caption{(Left): BMD (orange points), and test error (violet points) estimated for a two-layer fully connected network of width $10^3$, trained according to the adversarial initialization protocol described in section \ref{sec:adv} on 20K samples of the MNIST dataset. On the horizontal axis we vary the number of pretraining epochs, and plot the corresponding increase in the generalization error and the BMD of the model after the second learning stage. The points represent an average of 15 different realizations of the experiment. (Right): BMD (orange line) and counts (turquoise line) estimated for a two-layer fully connected network trained on the MNIST dataset using 20K train samples with 20\% label noise and tested on 5K samples. Counts represent the average amount of sign flips of random pixels of a correctly predicted test image that are necessary to fool the model to a wrong class label. The amount is averaged over all the correctly predicted test data samples. The resulting plot represents an average of 40 different realizations of the experiment. We observe that higher values of the BMD correspond to lower robustness of the model and vice versa.}
		\label{fig:adv_init1}
	\end{figure}
\end{center}

The adversarial initialization protocol works as follows. We train a two-layer fully connected network in two different phases: in the first phase, we push the network towards an adversarial initialization by pretraining the model with $100\%$ label noise for a fixed amount of epochs; in the second phase, we train the model on the original dataset, with no label noise, for $200$ epochs. The resulting plot, in Fig.~\ref{fig:adv_init1} (left panel), represents an average over $15$ different realizations of the experiment and shows the effect of the length of the pretraining phase on both generalization performance and BMD of the network. In agreement with our analysis, we observe a simultaneous increase of the two metrics when the adversarial initialization phase is longer and the network is driven towards worse generalization.

\subsection{BMD and Robustness Against Adversarial Attacks}

In this section, we analyze the connection between BMD of a model and its robustness to adversarial attacks. We consider a two-layer fully-connected network trained on MNIST with 10 classes. We define as our robustness measure the average count of sign flips of randomly chosen pixels, needed to change the model prediction on a test sample that was previously classified correctly. The lower the counts, the lower the robustness of the model. Varying the capacity of the model by varying the width of the hidden layer, we plot this robustness measure against the BMD of the model in Fig.~\ref{fig:adv_init1} (right panel). We observe that BMD and robustness strongly anti-correlate, with the peak in BMD coinciding with a minimum of robustness. 	

\subsection{Pixel-Wise Contributions to BMD}

\begin{figure}[H]
	\centering \includegraphics[scale=.55]{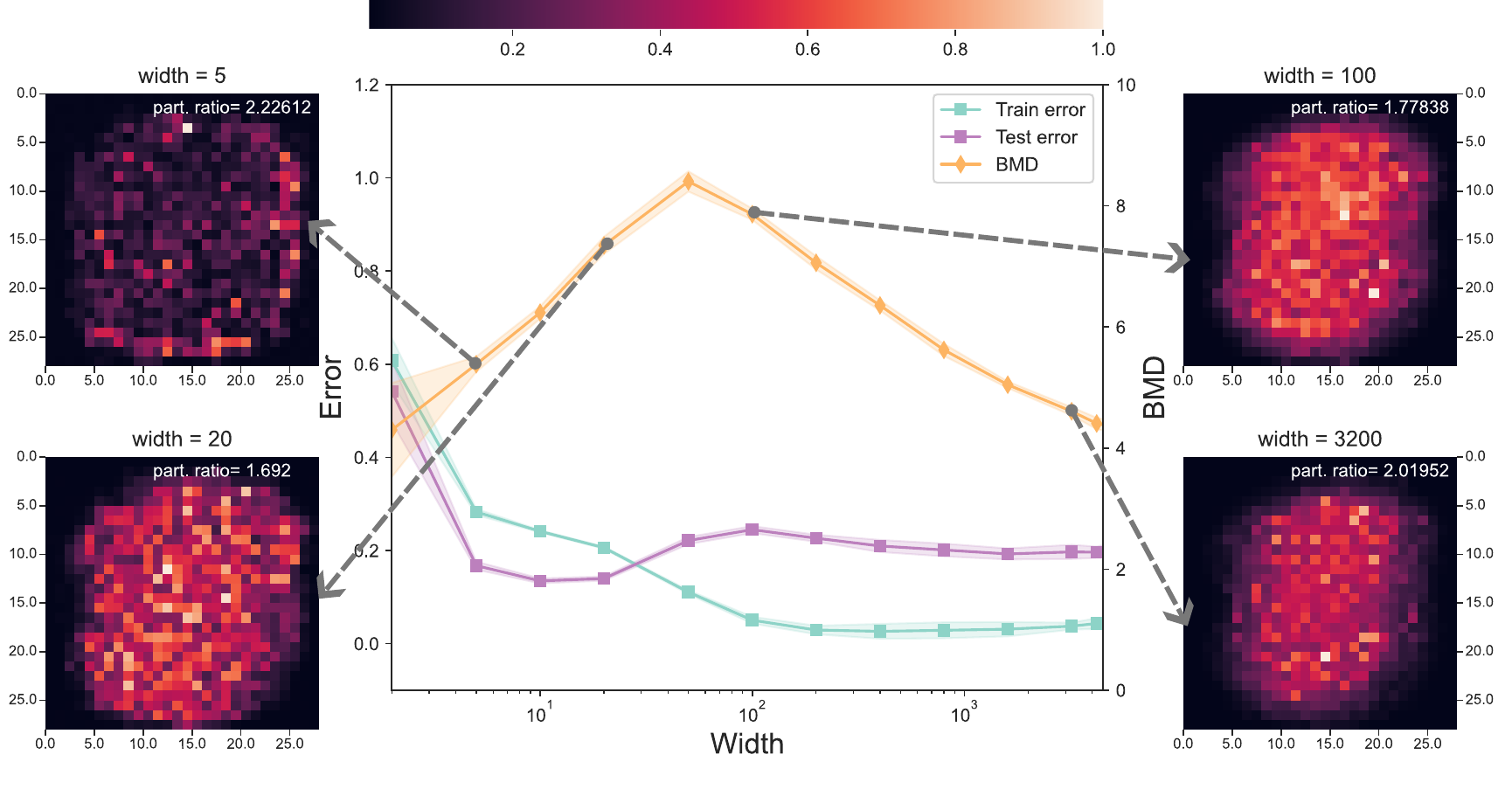}
	\caption{Heatmaps of the pixel contributions ($\tau^2_i$ for $1 \le i \le 784)$  estimated on the two-layer fully connected network trained on 20K samples of the MNIST dataset with 10 classes, 20\% label noise and normalized to lie within $[0, 1]$ interval. The (rescaled) participation ratio is defined as $n \times \frac{\sum_{i=1}^{n} \tau_i^2}{(\sum_{i=1}^n \tau_{i})^2}$. 
		After the rescaling, a participation ratio of $1$ indicates a uniform distribution of pixel contributions, while a value of $n$ indicates a distribution concentrated over a single pixel. The heatmaps correspond to the contributions estimated with respect to label $0$ for the models of different capacities (hidden layer dimensions) and represent only one seed, while the resulting curves on the plot represent an average over 20 different runs of the experiment.}
	\label{fig:heatmap}
\end{figure}

The MD as expressed in eq~\eqref{eq::MD_efficient_estimation} is proportional to a sum of contributions $\tau^2_i$ of single features indexed by $i$. Similar to~\cite{hahn2022mean}, we plot these contributions in Fig.~\ref{fig:heatmap} as a heatmap, where the bright spots indicate features that contribute strongly to the MD. We show four heatmaps, corresponding to different capacities and at different distances from the BMD peak, for a two-layer fully connected network trained on MNIST. 

Note that the colors are normalized to the $\left[0,1\right]$ range, so that very bright spots correspond to pixels that contribute to the BMD the most. It can be seen that for under-parametrized networks few pixels give the largest contribution to the BMD. Near the BMD peak, a large fraction of the pixels in the center of the image dominate the BMD, and for even larger capacities we again have fewer pixels with maximal values. This can be interpreted as the classifier losing ``focus'' at the interpolation point and paying attention to fewer patterns in the over-parametrized regime.

\subsection{Different Distributions for Estimating BMD} \label{sec::input_distribution}

\begin{figure}[H] 
	\centering \includegraphics[scale=.75]{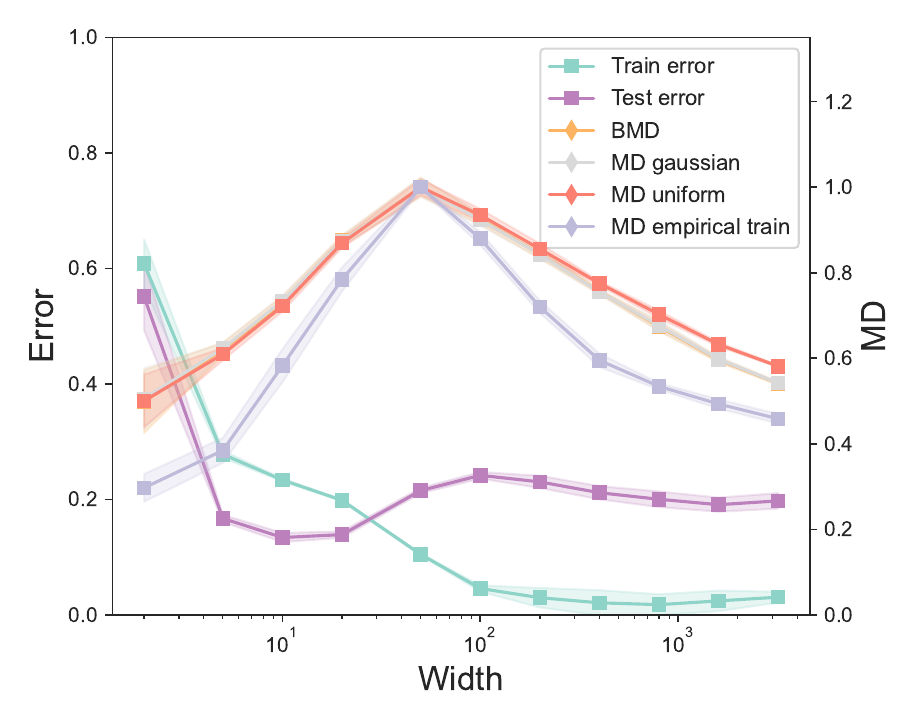}
	\caption{Mean dimensions estimated using Monte Carlo (eq. \ref{MD est}) w.r.t. different distributions of the two-layer fully connected network trained on the 20K of MNIST samples with 20\% label noise and tested on 5K samples. The MD values are normalized to lie within [0,1] interval. The choice of the distribution does not affect the location of the peak. Moreover, distributions, which first two moments coincide (e.g. binary uniform Unif$\{-1,1\}$ and Gaussian $\mathcal{N}(0,1)$) yield the same MD pattern. The resulting plot represents an average over 20 different runs of the experiment.}
	\label{Fig:diff_dist}
\end{figure}

In BMD estimates for the previous experiments, Eq.~\eqref{MD est}, we focused on the case of i.i.d. binary input features. In the RFM, however, we have shown analytically that there exists a universality for the MD when one considers separable input distributions with the same first and second moments. In the numerical experiments, we have also shown evidence that the BMD peak can still provide insights into the behavior of the neural network function on the training and test data, which follow very different input statistics. To explore in detail the role of the input statistics, and of the presence of correlations in the input features, we measure the MD by resampling the inputs from different distributions: in Fig.~\ref{Fig:diff_dist} we plot the normalized MD curves for features sampled from:
\begin{itemize}
	\item a uniform binary distribution (BMD).
	\item a standard normal (Gaussian) distribution $\mathcal{N}(0,1)$.
	\item a uniform distribution in the range $[-1, 1]$ .
	\item empirical distribution of the training data with random uniform resampling in the range $[-1, 1]$.
\end{itemize}

As one can see in Fig.~\ref{Fig:diff_dist}, the MD curves estimated with binary and Gaussian i.i.d. inputs, with matching moments, are identical. With the uniform distribution, the second moment is $1/3$ and this results in a slightly rescaled MD curve. Introducing correlations in the inputs, in the MD estimated over the training data distribution, the curve still shows a similar behavior, and importantly the peak is found at the same value.

\section{Discussion}

In this work, we analyzed the Boolean Mean Dimension as a tool for assessing the sensitivity of neural network functions. In the treatable setting of the Random Feature Model, we derived an exact characterization of the behavior of this metric as a function of the degree of overparameterization of the model. Notably, we found a strong correlation between the sharp increase of the BMD and the increase of the generalization error around the interpolation threshold. This finding indicates that as the neural network starts to overfit the noise in the data, the learned function becomes more sensitive to small perturbations of the input features. Importantly, while the double descent curve requires test data to be observed, the BMD can signal this type of failure mode by using information from the neural network alone.
The same phenomenology appears in more realistic scenarios with different architectures and datasets, where factors influencing double descent, like regularization and label noise, are also found to affect the BMD in similar fashion.
Furthermore, we demonstrated that the BMD is informative about the vulnerability of trained models to adversarial attacks, despite assuming an input distribution that is very different from that of the training dataset. 

Our study raises intriguing questions regarding the potential applications of BMD for regularization purposes. Another interesting future direction could be to investigate how comparing the BMDs achieved by a highly parametrized neural network trained on different datasets can help assess the effective dimensionality of the training data and the complexity of the discriminative tasks. Finally, it could be interesting to extend the study of the BMD in the RFM in the framework of a polynomial teacher model, recently analyzed in~\cite{aguirre2024random}.

\section*{Acknowledgements}
We thank Emanuele Borgonovo for the insightful discussions.
Enrico Malatesta and Luca Saglietti aknowledge the MUR-Prin 2022 funding, Prot. 20229T9EAT and 2022E3WYTY, financed by the EU (Next Generation EU).

\bibliographystyle{apalike}
\bibliography{abc.bib}

\newpage

\begin{center}
	{\huge \textbf{Supplemental Material}}
\end{center}
\appendix

\section{Proof of equation~\eqref{eq::MD_efficient_estimation}} \label{app::proof}

In order to prove the relation of the mean dimension of equation~\eqref{eq::MD_efficient_estimation} in the main text, we anticipate two Lemmas that are useful for the proof.
\begin{lemma}\label{Lemma1} For any $u \subseteq [n]$, the ANOVA coefficients satisfy the following relation:
	\begin{equation}
		\label{eq::ANOVA_relation}
		f_{u}(x) = \sum_{\substack{v \subseteq u}}(-1)^{|u|-|v|}\int f(\boldsymbol{x})p(\boldsymbol{x}_{\backslash v} | \boldsymbol{x}_{v})d\boldsymbol{x}_{\backslash v} = \sum_{\substack{v \subseteq u}}(-1)^{|u|-|v|}\mathbb{E}[f(\boldsymbol{x})|\boldsymbol{x}_v]
	\end{equation}
	\begin{proof}
		This can be seen by induction. The base of induction for the empty set and a singleton $u=\{i\}, \forall i \in n$ are respectively
		\begin{subequations}
			\begin{align}
				f_{\emptyset} &= \mathbb{E}[f(\boldsymbol{x})] \\
				f_{\{i\}} &= \mathbb{E}[f(\boldsymbol{x})|x_i] - \mathbb{E}[f(\boldsymbol{x})] = \mathbb{E}[f(\boldsymbol{x})|x_i] - f_{\emptyset}
			\end{align}
		\end{subequations}
		having used the ANOVA recursion in equation~\eqref{eq::ANOVA_recursion}. 
		Assuming equation~\eqref{eq::ANOVA_relation} holds for the sets up to the size $k$ we can show the induction step up to the set size $k+1$. Given a set $S \subseteq [n]$, $i \notin S$ and $|S|=k$ denoting $u =S\cup \{i\}$ we have:
		\begin{equation} 
			\begin{split}
				\label{eq::induction_step}
				f_{u}(\boldsymbol{x}_u) &= \mathbb{E}[f(\boldsymbol{x})|\boldsymbol{x}_{u}] - \sum_{v \subset u}f_v(\boldsymbol{x}_v) =  \mathbb{E}[f(\boldsymbol{x})|\boldsymbol{x}_{u}] - \sum_{v \subset u}\sum_{t \subseteq v}  (-1)^{|v|-|t|}\mathbb{E}[f(\boldsymbol{x})|\boldsymbol{x}_t]\\
				&= \mathbb{E}[f(\boldsymbol{x})|\boldsymbol{x}_{u}] - \sum_{\substack{t \subseteq S}} (-1)^{-|t|}\sum_{\substack{t \subseteq v \subset u}}
				(-1)^{|v|} \, \mathbb{E}[f(\boldsymbol{x})|\boldsymbol{x}_t] \,.
			\end{split}
		\end{equation}
		The summation over sets $t \subseteq v \subset u$ can be performed
		\begin{equation}
			\begin{split}
				\sum_{t \subseteq v \subset u}(-1)^{|v|} &= \sum_{k=|t|}^{|u|-1} (-1)^k {|u|-|t| \choose k-|t|} \\
				&= (-1)^{-|t|}\sum_{k=0}^{|u|-|t|-1}(-1)^k {|u|-|t| \choose k}=(-1)^{|u|+1} \,.
			\end{split}
		\end{equation}
		Inserting this back into~\eqref{eq::induction_step} we have finally
		\begin{equation}
			f_{u}(\boldsymbol{x}_u) = \mathbb{E}[f(\boldsymbol{x})|\boldsymbol{x}_{u}] + \sum_{t \subseteq S}(-1)^{|u|-|t|}\mathbb{E}[f(\boldsymbol{x})|\boldsymbol{x}_{t}] =  \sum_{v \subseteq u}(-1)^{|u|-|v|}\mathbb{E}[f(\boldsymbol{x})|\boldsymbol{x}_{v}]
		\end{equation}
	\end{proof}
\end{lemma}
\noindent Now given the previous Lemma \ref{Lemma1} we can show:
\begin{lemma}\label{Lemma2}
	We can write
	\begin{equation}
		f(\boldsymbol{x}) = \sum_{u \ni i}f_u(\boldsymbol{x}_u) + \mathbb{E} \left[f(\boldsymbol{x})| \boldsymbol{x}_{\backslash i}\right]
	\end{equation}
	or equivalently
	\begin{equation}
		\sum_{u \not \ni i}f_u( \boldsymbol{x}_u ) = \mathbb{E} \left[f(\boldsymbol{x})| \boldsymbol{x}_{\backslash i}\right]
	\end{equation}
	\begin{proof}
		To show that we consider:
		\begin{equation}
			\sum_{u \ni i}f_u(\boldsymbol{x}_u) = \sum_{u \ni i}\sum_{\substack{v \subseteq u}}(-1)^{|u|-|v|}\mathbb{E}\left[f(\boldsymbol{x})| \boldsymbol{x}_v\right]= \sum_{v} \left[\sum_{\substack{u \supseteq v, u\ni i}} (-1)^{|u|} \right](-1)^{-|v|}\mathbb{E}\left[f(\boldsymbol{x})| \boldsymbol{x}_v\right]
		\end{equation}
		The term in the squared brackets can be computed. We need to distinguish two cases, i.e. $i \in v$, and $i \notin v$. In the first case we get
		\begin{equation}
			\sum_{\substack{u \supseteq v, u\ni i}} (-1)^{|u|} = \sum_{k = |v|}^n (-1)^{k} \binom{n-|v|}{k-|v|} = \sum_{k = 0}^{n-|v|} (-1)^{k+|v|} \binom{n-|v|}{k} = (-1)^{|v|} \delta_{n, |v|}
		\end{equation}
		where $\delta_{i,j}$ is the Kronecker delta function. If $i \notin v$, instead
		\begin{equation}
			\sum_{\substack{u \supseteq v, u\ni i}} (-1)^{|u|} = \sum_{k = |v|+1}^n (-1)^{k} \binom{n-|v|-1}{k-|v|-1} = (-1)^{|v|+1}  \delta_{n, |v|+1}
		\end{equation}
		i.e. we get a non-zero result only if $v=[n]$ or $v=[n]\backslash \{i\}$. 
		Therefore we get
		\begin{equation*}
			\sum_{u \ni i}f_u(\boldsymbol{x}_u) = f(\boldsymbol{x}) - \mathbb{E}\left[ f(\boldsymbol{x}) | \boldsymbol{x}_{\backslash i}\right]
		\end{equation*}
	\end{proof}
\end{lemma}

\noindent In the following we will denote by $\boldsymbol{x}^{\oplus i}$ a vector $\boldsymbol{x}$ with a resampled $i_{th}$ coordinate. We are now ready to prove the following theorem:
\begin{theorem}
	The mean dimension can be written as
	\begin{equation}
		M_f = \sum\limits_{u \subseteq [n]} |u| \frac{\sigma^2_u}{\sigma^2} = \frac{\sum_{i=1}^n \tau_i^2}{\sigma^2}
	\end{equation}
	where
	\begin{equation}
		\tau_i^2 = \frac{1}{2} \int d \boldsymbol{x} \, d x_i' \, p(\boldsymbol{x}) \,  p(x_i'|\boldsymbol{x}_{\backslash i}) (f(\boldsymbol{x}) - f(\boldsymbol{x}^{\oplus i}))^2 \equiv \dfrac{1}{2}\mathbb{E}\left(f(\boldsymbol{x})-f(\boldsymbol{x}^{\oplus i})\right)^2
	\end{equation}
	\begin{proof}
		We can write the numerator of the mean dimension as
		\begin{equation}
			\sum\limits_{u \subseteq [n]} |u| \sigma^2_u = \sum_{k=1}^n k \sum_{\substack{u \subseteq [n]\\  \left| u \right| = k}} \sigma^2_u = \sum_{i=1}^n \sum_{\substack{u \subseteq [n]: i \in u}} \sigma^2_u \,.
		\end{equation}
		In the first equality we divided the summation over the sets into a double summation over the size $k$ of the set and a summation over the sets of fixed size $k$. In the second equality we have used the fact that the summation over $k$ can be interpreted as a summation over the indices $i$ of the variable $\boldsymbol{x}$; the inner sum can be therefore written as a summation over the sets (of any possible size) that contain the variable $i$ itself.
		Reminding that 
		\begin{equation}
			\sigma_u^2 \equiv 
			\int f_u^2(\boldsymbol{x}_u) \, p_u(\boldsymbol{x}_u) \, d \boldsymbol{x}_u = \mathbb{E} [f_u^2(\boldsymbol{x}_u)] \,,
		\end{equation}
		and using Lemma~\ref{Lemma2} and orthogonality of the coefficients of the ANOVA expansion, we have
		\begin{equation*}
			\begin{split}
				\tau_i^2 \equiv \dfrac{1}{2}\mathbb{E}\left(f(\boldsymbol{x})-f(\boldsymbol{x}^{\oplus i})\right)^2
				&= \dfrac{1}{2}\mathbb{E}\left(f(\boldsymbol{x})-f(\boldsymbol{x}^{\oplus i})\right)\left(\sum_{u \ni i}f_u(\boldsymbol{x}_u)-\sum_{u \ni i} f_u(\boldsymbol{x}_u^{\oplus i})\right)\\
				&= \mathbb{E}\left[ f(\boldsymbol{x})\sum_{u \ni i}f_{u}(\boldsymbol{x}_u) \right] - \mathbb{E} \left[f(\boldsymbol{x})\sum_{i \ni i}f_u(\boldsymbol{x}_u^{\oplus i}) \right] \\
				&= \mathbb{E} \left[\sum_{u \ni i}f^2_{u}(\boldsymbol{x}_u) \right] -
				\mathbb{E}\left[  f(\boldsymbol{x})\left(f(\boldsymbol{x}^{\oplus i}) - \mathbb{E} \left[ f(\boldsymbol{x}^{\oplus i}| \boldsymbol{x}_{\backslash i}) \right]\right)\right] \\
				&= \mathbb{E}\sum_{u \ni i}f^2_{u}(\boldsymbol{x}_u) - 0 = \sum_{u \ni i} \sigma^2_u \,.
			\end{split}
		\end{equation*}
	\end{proof}
\end{theorem}

\section{Replica computation of the Boolean Mean Dimension in the random feature model}~\label{app::MD_RFM}

\subsection{Free entropy}~\label{app::free_entropy}
We review here the replica calculations of the free entropy of the model, defined as
\begin{equation}
	\label{eq::free_entropy}
	\phi_\beta = \lim_{N \to \infty} \frac{1}{N} \mathbb{E}_{\mathcal{D}, F} \ln Z_\beta \,.
\end{equation}
This in turn will give the information necessary to compute the BMD. The average over the dataset in~\eqref{eq::free_entropy} can be performed using the replica trick
\begin{equation}
	\mathbb{E}_{\mathcal{D}, F} \ln Z_{\beta} = \lim\limits_{n \to 0} \frac{1}{n} \ln\left( \mathbb{E}_{\mathcal{D}, F} Z^n_\beta \right) \,.
\end{equation}
In the following we will consider $n$ as an integer and we will denote by $a, b$ as replica indices running from 1 to $n$. 

\paragraph*{Gaussian equivalence theorem}
In order to compute the average over the input patterns of $Z^n_\beta$, we will apply a central limit theorem~\cite{pennington2017nonlinear}, valid in the thermodynamic limit where $N$, $D$, $P$ go to infinity with fixed $\alpha \equiv \frac{P}{N}$ and $\alpha_D \equiv \frac{D}{N}$. In the statistical physics literature this central limit is often called Gaussian equivalence theorem (GET)~\cite{goldt2019modelling,loureiro2021capturing}. It can indeed be shown that the model is \emph{equivalent} to a Gaussian covariate model~\cite{mei2019generalization} and the following identification can be made
\begin{equation}
	\sigma\left( \frac{1}{\sqrt{D}} \sum_{k=1}^D F_{ki} \, x_k \right) = \kappa_0 + \frac{\kappa_1}{\sqrt{D}} \sum_{k=1}^D F_{ki} \, x_k + \kappa_\star \eta_i
\end{equation}
where $\eta_i$ is Gaussian noise with zero mean and unit variance and we have defined the following coefficients
\begin{subequations}
	\label{eq:effective_nonlinearity}
	\begin{align}
		\kappa_0 &= \int Dz \, \sigma(z) \\
		\kappa_1 &= \int Dz \, z \, \sigma(z) = \int Dz \, \sigma'(z) \\
		\kappa_2 &= \int Dz \, \sigma^2(z) \\
		\kappa_\star^2 &= \kappa_2 - \kappa_1^2 - \kappa_0^2
	\end{align}
\end{subequations}
with $Dz \equiv \frac{e^{-z^2/2}}{\sqrt{2\pi}} dz$.
In the following we will considered for simplicity $\sigma(\cdot)$ to be an odd activation function, so that in this case $\kappa_0 = 0$.

\paragraph*{Average over the dataset} Using GET, we therefore get
\begin{equation}
	\label{eq::part_func_after_average2}
	\begin{split}
		\mathbb{E}_{\mathcal{D}} Z^n_\beta &= \mathbb{E}_{\boldsymbol{w}^T} \int \prod_{a}  d\boldsymbol{w}^a \int \prod_{\mu} \frac{d u^\mu d \hat u^\mu}{2\pi} \prod_{\mu a} \frac{d \lambda^\mu_a d \hat \lambda^\mu_a}{2\pi}\, e^{-\beta \sum_{\mu a} \ell\left( \text{sign}(u^\mu) \lambda^\mu_a \right) -\frac{\beta \lambda}{2} \sum_{ia} (w_i^a)^2} \\
		& \times \prod_{\mu} e^{i u^\mu \hat u^\mu + i \sum_a \lambda_a^\mu \hat \lambda_a^\mu - \frac{(\hat u^\mu)^2}{2} -  \frac{1}{2} \sum_{ab} Q_{ab} \hat \lambda^\mu_a \hat \lambda^\mu_b -  \sum_a M_a \hat u^\mu \hat \lambda^\mu_a} \,.\\
		&= \int \prod_{a}  d\boldsymbol{w}^a \int \prod_{\mu} d u^\mu \prod_{\mu a} d \lambda^\mu_a\, e^{-\beta \sum_{\mu a} \ell\left( \text{sign}(u^\mu) \lambda^\mu_a \right) -\frac{\beta \lambda}{2} \sum_{ia} (w_i^a)^2} \prod_\mu\mathcal{N}(u^\mu, \lambda_a^\mu; \boldsymbol{0}, \Sigma_{ab})
	\end{split}
\end{equation}
where the correlation matrix of the $n+1$-dimensional multivariate Gaussian $\mathcal{N}$ is
\begin{equation}
	\Sigma_{ab} \equiv \begin{pmatrix}
		\rho & M_a \\
		M_a & Q_{ab} \,.
	\end{pmatrix}
\end{equation}
Here $\rho = \frac{1}{D} \sum_{k=1}^D (w_k^T)^2 = 1$, since the teacher has fixed norm. 
In the previous equation we have also denoted by $Q_{ab}$ and $M_a$ the following quantities
\begin{subequations}
	\label{eq::order_params_capital}
	\begin{align}
		\label{eq::order_params_capital_a}
		M_a &=  \kappa_1 \frac{1}{D} \sum_{k=1}^{D} s_k^a w_k^T \equiv \kappa_1 r_a\\
		\label{eq::order_params_capital_b}
		Q_{ab} &= \kappa_\star^2 \frac{1}{N} \sum_{i=1}^Nw_i^{a} w_i^b + \kappa_1^2  \frac{1}{D} \sum_{k=1}^{D} s_k^a s_k^b \equiv \kappa_\star^2 q_{ab} + \kappa_1^2  p_{ab}
	\end{align}
\end{subequations}
where we have defined the projected student weights in the space of the teacher $s_k^a$ as
\begin{equation}
	\label{eq::projected_weights}
	s_k^a \equiv \frac{1}{\sqrt{N}} \sum_{i=1}^{N} F_{ki} w_i^a \,.
\end{equation}

\paragraph*{Average over Gaussian Features}

We can enforce the definition of the projected weights in~\eqref{eq::projected_weights} using delta functions and their integral representation. It then becomes easy to perform the average over random Gaussian features. We get a terms of the following form
\begin{multline}
	\int \prod_{k a} \frac{d s_k^a d \hat s_k^a}{2\pi} \, e^{i \sum_{ka} s_k^a \hat s_k^a} \prod_{ki} \mathbb{E}_{F_{ki}} \left[ e^{-i \frac{F_{ki}}{\sqrt{N}} \sum_a \hat s_k^a w_i^a} \right] \\ = \int \prod_{k a} \frac{d s_k^a d \hat s_k^a}{2\pi} \, e^{i \sum_{ka} s_k^a \hat s_k^a - \frac{1}{2} \sum_{ab, k} \hat s_k^a \hat s_k^b \left( \frac{1}{N} \sum_i w_i^a w_i^b \right)} \,,
\end{multline}
which only depends on the $q_{ab}$ defined in~\eqref{eq::order_params_capital_b}. 

\paragraph*{Saddle point method} 
We can now impose the definitions of the order parameters
\begin{equation}
	q_{ab} \equiv \frac{1}{N} \sum_i w_i^a w_i^b\,, \qquad p_{ab} \equiv \frac{1}{D} \sum_k s_k^a s_k^b\,, \qquad r_a \equiv \frac{1}{D} \sum_k s_k^a w_k^T\,.
\end{equation}
Denoting by $\left \langle \cdot \right \rangle_{\mathcal{D}, F}$ the average over both patterns and random features, the final result reads
\begin{equation}
	\mathbb{E}_{\mathcal{D}, F} Z^n_\beta = \int \prod_{a \le b} \frac{d q_{ab} d \hat q_{ab}}{2 \pi} \prod_{a \le b} \frac{d p_{ab} d \hat p_{ab}}{2 \pi} \prod_a \frac{d r_{a} d \hat r_{a}}{2 \pi} \, e^{N \phi_\beta^{(n)}}
\end{equation}
where 
\begin{subequations}
	\label{eq::termsSP}
	\begin{align}
		\phi^{(n)}_\beta &= - \frac{1}{2}\sum_{a b} q_{ab} \hat q_{ab} - \frac{\alpha_D}{2} \sum_{ab} p_{ab} \hat p_{ab} - \alpha_D \sum_{a} r_a \hat r_a + G_{SS} + \alpha_D G_{SE} + \alpha G_{E} \\
		G_{SS} &= \ln \int \prod_a dw_a \, e^{ \, \frac{1}{2}\sum_{ab} \hat q_{ab} w_a w_b - \frac{\beta \lambda}{2} \sum_a w_a^2}\\
		\label{eq::termsSP_GSE}
		G_{SE} &= \ln \int \prod_a \frac{d s_a d \hat s_a}{2\pi} \, e^{i \sum_a s_a \hat s_a + \sum_a \hat r_a s_a + \frac{1}{2}\sum_{ab} \hat p_{ab} s_a s_b - \frac{1}{2} \sum_{ab} q_{ab} \hat s_a \hat s_b} \\
		G_{E} &= \ln \int \prod_a \frac{d \lambda_a d \hat \lambda_a}{2\pi} \frac{d u d \hat u}{2\pi} \, e^{i u \hat u + i \sum_a \lambda_a \hat \lambda_a -\beta \sum_a \ell\left(\text{sign}(u) \lambda_a \right) - \frac{\hat u^2}{2} - \frac{1}{2} \sum_{ab} Q_{ab} \hat \lambda_a \hat \lambda_b - \hat u \sum_a M_a \hat \lambda_a }
	\end{align}
\end{subequations}
and $M_a$, $Q_{ab}$ are defined in terms of $q_{ab}$, $p_{ab}$, $r_a$ in~\eqref{eq::order_params_capital}. Notice that the integrals inside the ``entropic'' $G_S$ and the $G_{SE}$ tersm can be solved analytically by using multivariate Gaussian integrals identities.

\paragraph*{Replica Symmetric Ansatz} 
We focus on the saddle points that preserve the symmetry between the exchange of replicas (RS)
\begin{subequations}
	\begin{align}
		& & r_a &= r\,, &\hat{r}_a &= \hat{r}\,, & &&\forall a &\in [n] \\
		q_{aa} &= q_d\,, &\hat{q}_{aa} &= - \hat{q}_d\,, & p_{aa} &= p_d\,, &\hat{p}_{aa} &= - \hat{p}_d \,, &\forall a &\in [n] \\
		q_{ab} &= q\,, &\hat{q}_{ab} &= \hat{q}\,, & p_{ab} &= p\,, &\hat{p}_{ab} &= \hat{p} \,, & 1 \le a &\le b  \le n
	\end{align}
\end{subequations}
Denoting with 
\begin{subequations}
	\begin{align}
		M &\equiv \kappa_1 r \,, \\ 
		Q &\equiv \kappa_\star^2 q + \kappa_1^2 p\,, \\ 
		Q_d & \equiv \kappa_\star^2 q_d + \kappa_1^2 p_d
	\end{align}
\end{subequations}
we finally get in the small $n$ limit the free entropy reads
\begin{equation}
	\phi_\beta \equiv \lim\limits_{n \to 0} \frac{\phi^{(n)}_\beta}{n} = \frac{1}{2}(q_d \hat q_d + q \hat q) + \frac{\alpha_D}{2} \left( p_d \hat p_d + p \hat p \right) - \alpha_D r \hat r + \mathcal{G}_{SS} + \alpha_D \mathcal{G}_{SE} + \alpha \mathcal{G}_{E} \,,
\end{equation}
where
\begin{subequations}
	\begin{align}
		\mathcal{G}_{SS} &\equiv \lim\limits_{n \to 0} \frac{G_{SS}}{n} = \frac{1}{2} \ln\left( \frac{2\pi}{\hat q_d + \hat q + \beta \lambda} \right) + \frac{\hat q}{2(\hat q_d + \hat q + \beta \lambda)}\\
		\mathcal{G}_{SE} &\equiv \lim\limits_{n \to 0} \frac{G_{SE}}{n} = - \frac{q}{2(q_d-q)} - \frac{1}{2} \ln \left[ 1 + (\hat p + \hat p_d) (q_d-q) \right] + \frac{1}{2} \frac{(\hat p + \hat r^2)(q_d-q) + \frac{q}{q_d-q}}{1+ (\hat p + \hat p_d)(q_d-q)} \\
		\mathcal{G}_{E} &\equiv \lim\limits_{n \to 0} \frac{G_{E}}{n} = 2 \int Dz_0 \, H \left( - \frac{M z_0}{\sqrt{Q - M^2}} \right) \ln \int Dz_1 \, e^{-\beta \ell\left(  \sqrt{Q} z_0 + \sqrt{Q_d - Q} z_1\right)}
	\end{align}
\end{subequations}
and $H(x) = \int_x^\infty Dz = \frac{1}{2} \text{Erfc}\left( \frac{x}{\sqrt{2}} \right)$. The values of the 10 order parameters, namely $q$, $\hat q$, $q_d$, $\hat{q}_d$, $p$, $\hat{p}$, $p_d$, $\hat{p}_d$, $r$, $\hat{r}$ must be found self-consistently by solving the saddle point equations obtained by differentiating $\phi$.

\paragraph*{Large $\beta$ limit} 
We now restrict the discussion to the case of the convex loss functions as the MSE and the CE loss cited in the main text~\eqref{eq::loss_functions}. In this case, we have to impose the following scalings
\begin{subequations}
	\begin{align}
		\hat r &\to \beta \hat r\,, & & \\
		q &= q_d - \frac{\delta q}{\beta}\,, & p &= p_d - \frac{\delta p}{\beta} \,, \\
		\hat q &= \beta^2 \delta \hat q + \frac{\beta}{2} \delta \hat Q\,, & \hat q_d &= -\beta^2 \delta \hat q + \frac{\beta}{2} \delta \hat Q\,,\\
		\hat p &= \beta^2 \delta \hat p + \frac{\beta}{2} \delta \hat P \,, & \hat p_d &= -\beta^2 \delta \hat p + \frac{\beta}{2} \delta \hat P\,.
	\end{align}
\end{subequations}
We have that the free energy which was defined in the main text in equation~\eqref{eq::free_energy} is
\begin{equation}
	\label{eq::free_energy_replica}
	-f = \lim\limits_{\beta \to \infty} \frac{\phi_\beta}{\beta} = \frac{1}{2}\left(q_d \delta \hat Q - \delta q \delta \hat q\right) + \frac{\alpha_D}{2} \left(p_d \delta \hat P -  \delta p \delta \hat p \right) - \alpha_D r \hat r + \mathcal{G}_{SS} + \alpha_D \mathcal{G}_{SE} + \alpha \mathcal{G}_{E} \,,
\end{equation}
where
\begin{subequations}
	\begin{align}
		\mathcal{G}_{SS} &= \frac{\delta \hat q}{2(\delta \hat Q + \lambda)}\\
		\mathcal{G}_{SE} & = - \frac{q_d}{2 \delta q} + \frac{1}{2} \frac{(\delta \hat p + \hat r^2) \, \delta q + \frac{q_d}{\delta q}}{1+ \delta \hat P \, \delta q} \\
		\mathcal{G}_{E} &= 2 \int Dz_0 \, H \left( - \frac{M z_0}{\sqrt{Q_d - M^2}} \right) \max_{z_1} \left[-\frac{z_1^2}{2} - \ell(\sqrt{Q_d} z_0 + \sqrt{\delta Q} z_1) \right] \,.
	\end{align}
\end{subequations}
We have also denoted by $\delta Q = \kappa_\star^2 \delta q + \kappa_1^2 \delta p$. Again the order parameters $\delta q$, $\delta \hat q$, $q_d$, $\delta \hat{Q}$, $\delta p$, $\delta \hat{p}$, $p_d$, $\delta \hat{P}$, $r$, $\hat{r}$ must be found self-consistently by solving the saddle point equations obtained by differentiating $f$.

\paragraph*{Physical observables of interest}

As shown in multiple papers, see e.g.~\cite{gerace2020generalisation,baldassi2022learning} the generalization error can be obtained by computing
\begin{equation}
	\epsilon_g = \frac{1}{\pi} \arccos\left( \frac{M}{\sqrt{Q_d}} \right) \,.
\end{equation}
The training loss can be computed by a derivative with respect to $\beta$
\begin{equation}
	\ell_t = \lim\limits_{\beta \to \infty} \frac{\partial (\beta f)}{\partial \beta} = 2 \int Dz_0 \, H \left( - \frac{M z_0}{\sqrt{Q_d - M^2}} \right)  \ell\left(  \sqrt{Q_d} z_0 + \sqrt{\delta Q} z_1^\star\right)
\end{equation}
with 
\begin{equation}
	z_1^\star = \argmax_{z_1} \left[-\frac{z_1^2}{2} - \ell(\sqrt{Q_d} z_0 + \sqrt{\delta Q} z_1) \right]
\end{equation}
The test loss can be computed as follows~\cite{baldassi2022learning}
\begin{equation}
	\begin{split}
		\ell_g &\equiv \langle \mathbb{E}_{\boldsymbol{\xi}^\star} \ell\left( y^\star \hat y^\star \right) \rangle \\
		&= \int \frac{d u d \hat u}{2\pi} \frac{d \lambda d \hat \lambda}{2\pi}\, \ell\left( \text{sign}(u) \lambda \right) e^{i u \hat u + 
			i \lambda \hat \lambda	- \frac{\hat u^2}{2} -  \frac{1}{2} Q_{d} \hat \lambda^2 -  M \hat u \hat \lambda} 
		\\ 
		&= 
		2 \int Dv \, \ell\left(- \sqrt{Q_d} v \right) H \left(- \frac{M v}{\sqrt{Q_d - M^2} } \right)\,.
	\end{split}
\end{equation}
where $\boldsymbol{\xi}^\star$ represents a new extracted pattern, $y^\star$ its corresponding label and $\hat{y}^\star$ the prediction of the model. 

\subsection{Analytical determination of the BMD}~\label{app::BMD}

In the following subsections we will show the derivation of the annealed averages on the inputs of the numerator and denominator of the BMD. 

\subsubsection{Annealed average of the denominator of the BMD}
The denominator in the definition of the BMD is easy to analyze. Applying GET, we obtain
\begin{equation}
	\begin{split}
		\left \langle \hat{y}^2(\boldsymbol{w}; \boldsymbol{x}) \right \rangle  &= \left \langle  \frac{1}{N} \sum_{i,j} w_i w_j \, \sigma\left( \frac{1}{\sqrt{D}} \sum_{k=1}^D F_{ki} \, x_k \right) \sigma\left( \frac{1}{\sqrt{D}} \sum_{k=1}^D F_{kj} \, x_k \right) \right \rangle \\
		&= \left \langle  \frac{1}{N} \sum_{i,j} w_i w_j \,\left( \kappa_0 +  \frac{\kappa_1}{\sqrt{D}} \sum_{k=1}^D F_{ki} \, x_k + \kappa_\star \eta_i \right) \left(\kappa_0 +  \frac{\kappa_1}{\sqrt{D}} \sum_{k=1}^D F_{kj} \, x_k + \kappa_\star \eta_j \right) \right \rangle \\
		&= \left( \frac{\kappa_0}{\sqrt{N}} \sum_i w_i \right)^2 + \frac{\kappa_1^2}{D} \sum_k \left( \frac{1}{\sqrt{N}} \sum_i F_{ki} w_i \right) \left( \frac{1}{\sqrt{N}} \sum_j F_{kj} w_j \right) + \frac{\kappa_\star^2}{N}\sum_i w_i^2 = Q_d\\
	\end{split}
\end{equation}
The average squared is instead simply given by
\begin{equation}
	\left \langle \hat{y}(\boldsymbol{w}; \boldsymbol{x}) \right \rangle^2 =  \left( \frac{\kappa_0}{\sqrt{N}} \sum_i w_i \right)^2
\end{equation}
Therefore the denominator in the BMD reads
\begin{equation}
	\left \langle \hat{y}^2(\boldsymbol{w}; \boldsymbol{x}) \right \rangle  - \left \langle \hat{y}(\boldsymbol{w}; \boldsymbol{x}) \right \rangle^2 = \frac{\kappa_1^2}{D} \sum_k \left( \frac{1}{\sqrt{N}} \sum_i F_{ki} w_i \right) \left( \frac{1}{\sqrt{N}} \sum_j F_{kj} w_j \right) + \frac{\kappa_\star^2}{N}\sum_i w_i^2 = Q_d
\end{equation}
where $Q_d$ is the overlap obtained by the RS saddle point equations sketched in the previous section.

\subsubsection{Annealed average of the numerator of the BMD}
More work has to be done to compute the numerator of the BMD. We can write it as
\begin{equation}
	\begin{split}
		&\sum_{k'=1}^D \left \langle (\hat{y}(\boldsymbol{w}; \boldsymbol{x}) - \hat{y}(\boldsymbol{w}; \boldsymbol{x}^{\oplus k'}))^2 \right \rangle =  \\
		&= \sum_{k'=1}^D \left \langle \left( \frac{1}{\sqrt{N}} \sum_{i=1}^N w_i \, \sigma\left( \frac{1}{\sqrt{D}} \sum_{k=1}^D F_{ki} \, x_k \right) -  \frac{1}{\sqrt{N}} \sum_{i=1}^N w_i \, \sigma\left( \frac{1}{\sqrt{D}} \sum_{k\ne k'} F_{ki} \, x_k + \frac{F_{k'i} \tilde{x}_{k'}}{\sqrt{D}} \right)   \right)^2 \right\rangle \\
		&= \sum_{k'=1}^D \left \langle \left( \frac{1}{\sqrt{N}} \sum_{i=1}^N w_i \, \sigma\left( \frac{1}{\sqrt{D}} \sum_{k=1}^D F_{ki} \, x_k \right) -  \frac{1}{\sqrt{N}} \sum_{i=1}^N w_i \, \sigma\left( \frac{1}{\sqrt{D}} \sum_{k=1}^D F_{ki} \, x_k - \frac{F_{k'i} (x_{k'}-\tilde{x}_{k'})}{\sqrt{D}} \right) \right)^2 \right\rangle
	\end{split}
\end{equation}
Taylor expanding the second term we have
\begin{equation}
	\begin{split}
		&\frac{1}{2}\sum_{k'=1}^D \left \langle (\hat{y}(\boldsymbol{w}; \boldsymbol{x}) - \hat{y}(\boldsymbol{w}; \boldsymbol{x}^{\oplus k'}))^2 \right \rangle \\
		&= \frac{1}{2} \sum_{k'=1}^D \left \langle \left( \frac{1}{\sqrt{N}} \sum_{i=1}^N w_i \, \sigma'\left( \frac{1}{\sqrt{D}} \sum_{k=1}^D F_{ki} \, x_k \right)  \frac{F_{k'i} (x_{k'}-\tilde{x}_{k'})}{\sqrt{D}} \right)^2 \right\rangle \\
		&= \frac{1}{2} \sum_{k'=1}^D \left \langle  \frac{1}{N} \sum_{i,j} w_i w_j \, \sigma'\left( \frac{1}{\sqrt{D}} \sum_{k=1}^D F_{ki} \, x_k \right) \sigma'\left( \frac{1}{\sqrt{D}} \sum_{k=1}^D F_{kj} \, x_k \right)  \frac{F_{k'i} F_{k'j} (x_{k'}-\tilde{x}_{k'})^2}{D}  \right\rangle \,.
	\end{split}
\end{equation}
We then apply the GET and we take the average over the inputs getting
\begin{equation}
	\begin{split}
		& \frac{1}{2}\left\langle \sigma'\left( \frac{1}{\sqrt{D}} \sum_{k=1}^D F_{ki} \, x_k \right) \sigma'\left( \frac{1}{\sqrt{D}} \sum_{k=1}^D F_{kj} \, x_k \right)  (x_{k'}-\tilde{x}_{k'})^2  \right\rangle \\
		&= \frac{1}{2}\left\langle \left[\bar{\kappa}_0 + \frac{\bar{\kappa}_1}{\sqrt{D}} \sum_{k=1}^D F_{ki} \, x_k  + \bar{\kappa}_\star \eta_i \right] \left[\bar{\kappa}_0 + \frac{\bar{\kappa}_1}{\sqrt{D}} \sum_{k=1}^D F_{kj} \, x_k  + \bar{\kappa}_\star \eta_j \right]  (x_{k'}-\tilde{x}_{k'})^2  \right\rangle \\
		&= \bar{\kappa}_0^2 + \bar{\kappa}_\star^2 \delta_{ij} + \frac{\bar{\kappa}_1^2}{2D} \sum_{k=1}^D \sum_{k''=1}^D F_{ki} \, F_{k''j} \, \left\langle x_k x_{k''} (x_{k'}-\tilde{x}_{k'})^2 \right\rangle \\
		&= \bar{\kappa}_0^2 + \bar{\kappa}_\star^2 \delta_{ij} + \frac{\bar{\kappa}_1^2}{D} \left( 2 F_{k'i} \, F_{k'j} + \sum_{k} F_{ki} \, F_{kj} \right) = \bar{\kappa}_0^2 + \bar{\kappa}_\star^2 \delta_{ij} + \frac{\bar{\kappa}_1^2}{D} \sum_{k} F_{ki} \, F_{kj} \,.
	\end{split}
\end{equation}
where we have denoted for simplicity by $\bar{\kappa}_0$, $\bar{\kappa}_1$, $\bar{\kappa}_2$, $\bar{\kappa}_\star$ the coefficients in equation~\eqref{eq:effective_nonlinearity} for $\sigma'$. 
Notice that in all the steps that we have done to arrive performing the average over the inputs we have not used anywhere the binary nature of the inputs. It is therefore easy to see that we would have obtained the same result if we had chosen a probability distribution on the inputs with the same first two moments (e.g., a standard normal distribution).
Inserting this back into the previous equation, we get
\begin{equation}
	\begin{split}
		&\frac{1}{2}\sum_{k'=1}^D \left \langle (\hat{y}(\boldsymbol{w}; \boldsymbol{x}) - \hat{y}(\boldsymbol{w}; \boldsymbol{x}^{\oplus k'}))^2 \right \rangle \\
		&= \sum_{k'=1}^D  \frac{1}{N} \sum_{ij} w_i w_j \, \left[ \bar{\kappa}_0^2 + \bar{\kappa}_\star^2 \delta_{ij} + \frac{\bar{\kappa}_1^2}{D} \sum_{k} F_{ki} \, F_{kj} \right]  \frac{F_{k'i} F_{k'j}}{D} \\
		&= \frac{\bar{\kappa}_0^2 }{N} \sum_{ij} \Omega_{ij} w_i w_j  + \frac{\bar{\kappa}_\star^2 }{N} \sum_{i} \Omega_{ii} w_i^2  + \frac{\bar{\kappa}_1^2 }{N} \sum_{ij} \Omega^2_{ij} w_i w_j = \frac{1}{N} \sum_{ij} \bar{\Psi}_{ij} w_i w_j
	\end{split}
\end{equation}
where we have introduced
\begin{subequations}
	\begin{align}
		\bar{\Psi}_{ij} &\equiv \bar{\kappa}_\star^2 \, \Omega_{ii} \delta_{ij} + \bar{\kappa}_0^2 \, \Omega_{ij} + \bar{\kappa}_1^2 \, \Omega^2_{ij} \,, \\
		\Omega_{ij} &\equiv \frac{1}{D} \sum_{k=1}^D F_{ki} F_{kj} \,.
	\end{align}
\end{subequations}
Therefore the mean dimension for a generic non-linearity $\sigma$ is
\begin{equation}
	M_f(\boldsymbol{w}) = \frac{\frac{1}{N} \sum_{ij} \bar{\Psi}_{ij} w_i w_j}{Q_d}
\end{equation}

\subsubsection{BMD for an odd non-linearity $\sigma$} \label{app::BMD_anal_odd_activation}
If we assume the activation function to be odd we have $\bar{\kappa}_1=0$, and $\bar{\kappa}_0=\kappa_1$. Therefore we can write the BMD in terms of the order parameters only 
\begin{equation}
	\label{eq::BMD_FINAL}
	M_f \equiv \left \langle M_f(\boldsymbol{w}) \right\rangle_{\boldsymbol{w}} = \frac{Q_d + \left(\bar{\kappa}_\star^2 - \kappa_\star^2\right) q_d }{Q_d}
\end{equation}
where
\begin{equation}
	\bar{\kappa}_\star^2 - \kappa_\star^2 
	= \int Dz \left(\sigma'(z)\right)^2 - \int Dz \, \sigma^2(z) = \bar \kappa_2 - \kappa_2
\end{equation}
\begin{figure}[H]
	\captionsetup{justification=raggedright, singlelinecheck=false}
	\centering \includegraphics[width=0.5\textwidth]{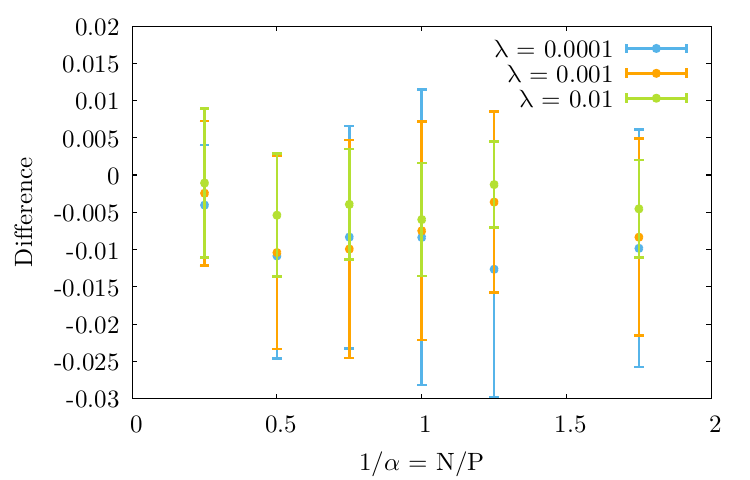}
	\caption{Difference between the BMD estimated using Monte Carlo method and using equation~\eqref{eq::BMD_FINAL}.}
	\label{fig:MD_MC_VS_ANAL}
\end{figure}
Notice that equation~\eqref{eq::BMD_FINAL} can be used as an alternative (very efficient) way of computing the BMD, without using Monte Carlo method. We show in Fig.~\ref{fig:MD_MC_VS_ANAL} how the difference between the BMD estimated using Monte Carlo and the one using equation~\eqref{eq::BMD_FINAL} showing good agreement.

In the limit of $1/\alpha\to 0$, at fixed $\alpha_T = P/D$ (i.e. $N\to0$), the solution to the saddle point equations show that $p_d \to q_d$; in this limit the BMD goes to
\begin{equation}
	M_f = 1 + \frac{\bar{\kappa}_\star^2 - \kappa_\star^2}{\kappa_\star^2 + \kappa_1^2} = \frac{\bar{\kappa}_2}{\kappa_2}\,. 
\end{equation}
For example for $\sigma(x) = \tanh(x)$ activation we get $M_f = 1.1778$ as it is displayed in Fig.~\ref{fig:MD_analytic}.

\section{Double peak behavior of the BMD}~\label{app::BMD_two_peaks}

As can be seen in Fig.~\ref{fig:MD_triple_descent}, if $\alpha_T$ is sufficiently large the BMD can display a peak in addition to the one located at the interpolation threshold ($N=P$ for the MSE loss). This secondary peak is located at $\alpha=\alpha_T$, i.e. when the number of parameters is equal to the input dimension $N = D$. This peak is not present in the generalization. 
\begin{figure}[H]
	\captionsetup{justification=raggedright, singlelinecheck=false}
	\includegraphics[width=0.5\textwidth]{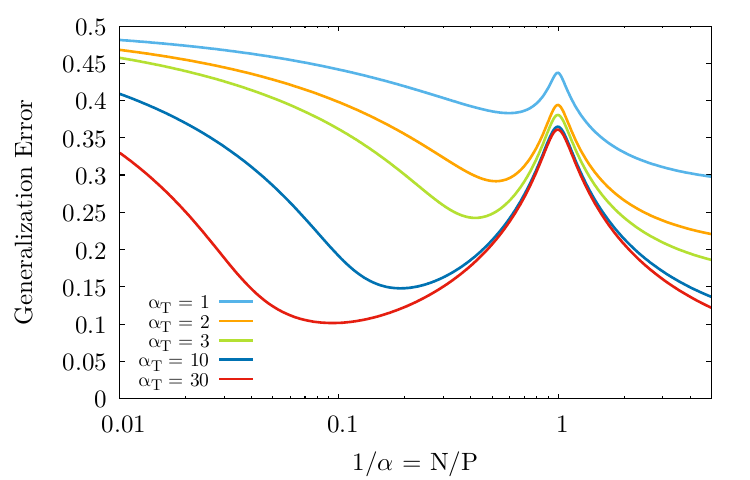}
	\includegraphics[width=0.5\textwidth]{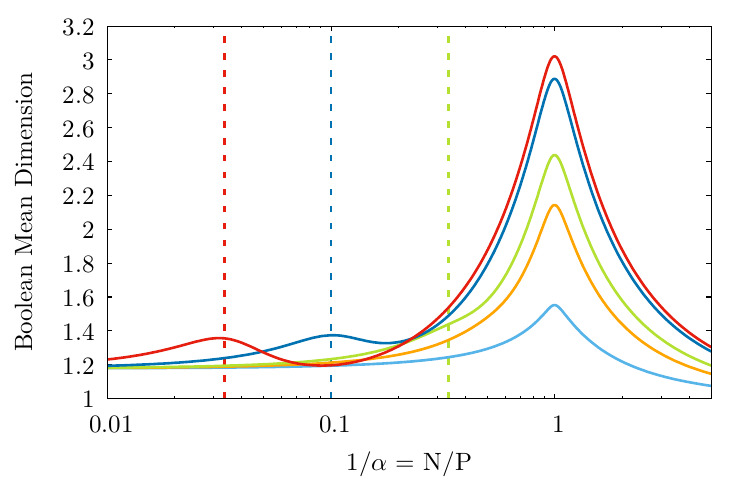}
	\caption{Generalization error (left) and BMD (right) as a function of $1/\alpha$ for $\lambda = 10^{-4}$ and $\sigma = \tanh$, for several values of $\alpha_T = P/D$. The loss used is the MSE. For low values of $\alpha_T$ the BMD displays a double descent behavior as showed in the main text. Increasing $\alpha_T$, contrary to generalization, the BMD shows a triple descent behavior: a secondary peak in the BMD appears at $\alpha = \alpha_T$, i.e. $N=D$ (dashed vertical lines).}
	\label{fig:MD_triple_descent}
\end{figure}
We remark here that this behavior observed in the BMD is reminescent of the triple descent behaviour observed in~\cite{Biroli2020triple}, but is nonetheless different in nature. Indeed, in~\cite{Biroli2020triple} the triple descent was observed in the test loss, when fixing $N$ and $D$ (i.e. $\alpha_D = D/N$) and changing $P$\footnote{this is different from our setting where we fix $P$ and $D$ i.e. $\alpha_T = P/D$ and change $N$.}; the authors observe a peak in the test loss when $P=D$ in addition to the ``classical'' double descent peak when $P=N$. This ``secondary'' peak can be observed only if the activation function is linear $\sigma(x) = x$ or if the labels are corrupted by Gaussian noise $\zeta^\mu \sim \mathcal{N}(0,1)$
\begin{equation}
	y^\mu = \text{sign}\left( \frac{1}{\sqrt{D}} \sum_k w_k^T \xi_k^\mu + \sqrt{\Delta} \zeta^\mu  \right) \,.
\end{equation}
where $\Delta$ is a non-negative parameter modulating the noise intensity. It is easy to show that the only term to change in the free energy~\eqref{eq::free_energy_replica} because of the noise is the energetic term which is modified as
\begin{equation}
	\mathcal{G}_{E} = 2 \int Dz_0 \, H \left( - \frac{M z_0}{\sqrt{Q_d - M^2 + \Delta}} \right) \max_{z_1} \left[-\frac{z_1^2}{2} - \ell(\sqrt{Q_d} z_0 + \sqrt{\delta Q} z_1) \right] \,.
\end{equation}
\begin{figure}[H]
	\captionsetup{justification=raggedright, singlelinecheck=false}
	\includegraphics[width=0.5\textwidth]{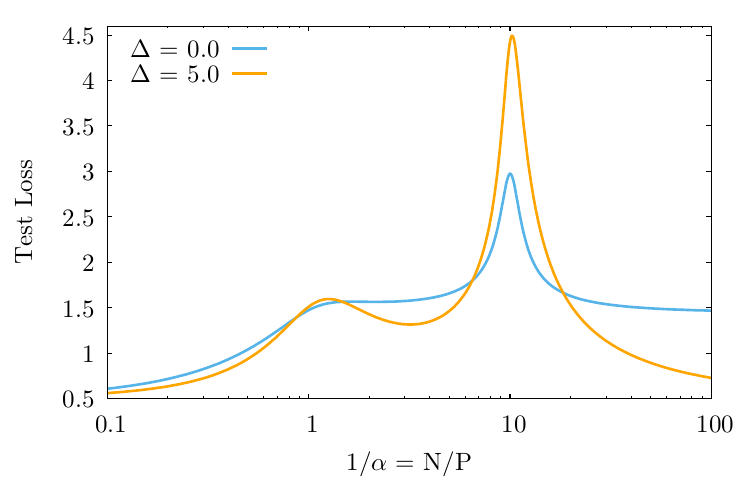}
	\includegraphics[width=0.5\textwidth]{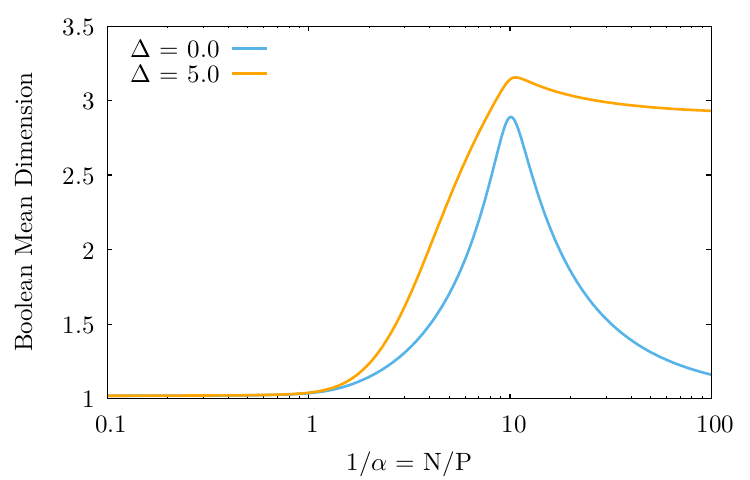}
	\caption{Test loss (left) and BMD (right) as a function of $1/\alpha$ for fixed $\alpha_D = 0.1$, $\lambda = 10^{-4}$, $\sigma = \tanh$ and for 2 values of the noise in the labels $\Delta =0, 5$. The loss used is the MSE. Even if the test loss displays a secondary peak at $P = D$, the BMD does not.}
	\label{fig:MD_triple_descent}
\end{figure}
We show in Fig.~\ref{fig:MD_triple_descent} that even if the test loss has a secondary peak when $\Delta>0$ at $P=D$, this peak is not present in the BMD. 

\begin{figure}[H]
	\captionsetup{justification=raggedright, singlelinecheck=false}
	\includegraphics[width=0.5\textwidth]{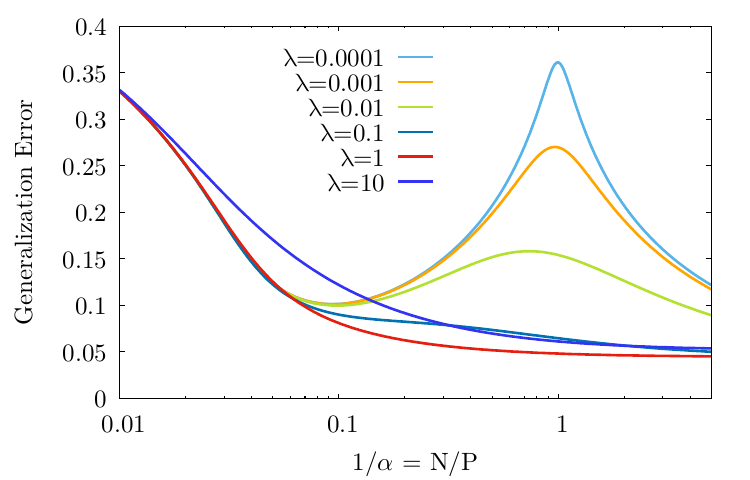}
	\includegraphics[width=0.5\textwidth]{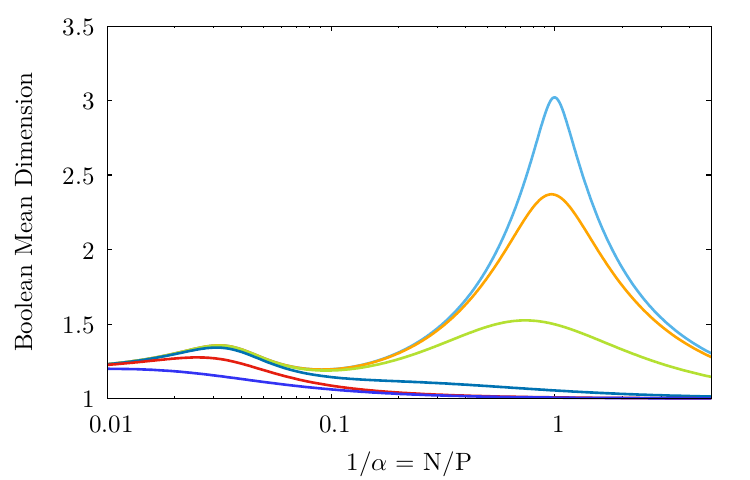}
	\caption{Generalization error (left) and BMD (right) as a function of $1/\alpha$ for $\alpha_T = 30$ and $\sigma = \tanh$ for several values of the regularization $\lambda$. The loss used is the MSE. Increasing the regularization the peak corresponding to $P=N$ disappears before the one located at $N=D$.}
	\label{fig:Avoiding_MD_triple_descent}
\end{figure}

In Fig.~\ref{fig:Avoiding_MD_triple_descent}, we show how the regularization $\lambda$ can not only attenuate the ``primary'' peak at $P = N$, but also make the secondary peak disappear at $N = D$.

\subsection{Explanation the secondary peak of the BMD around $N=D$}

The root of this phenomenon can be found in the behavior of the spectrum of the covariance matrix $\Omega = F^T F/D$. To see this, we first consider a simplified setting where the phenomenon can be easily analytically traced.
Consider a linear regression in the RFM with linear teacher. 
Calling $X_p\in\mathbb{R}^{P \times N}=\sigma(X F /\sqrt{D})$ the projected inputs of the RFM, with $X\in\mathbb{R}^{P \times D}$,  and $F\in\mathbb{R}^{D \times N}$, the Gaussian Equivalence implies that the learning problem is equivalent to a linear regression with data:
\begin{equation}
	X_p^{GET} = \kappa_1 X F / \sqrt{D} + \kappa_\star Z
\end{equation}
with $Z\in\mathrm{R}^{P\times N}$ and $Z_{ij}\sim\mathcal{N}(0,1)$.
The labels $Y\in \mathbb{R}^P$ are given by a linear teacher $w_T\in\mathbb{R}^D$:
\begin{equation}
	Y = X w_T
\end{equation}
The ordinary least square (OLS) estimator gives a closed form solution for the trained weights:
\begin{equation}
	\begin{split}
		w &= (X_p^T X_p / N + \lambda \mathbb{I})^{-1} X_p^T Y / \sqrt{N} \\
		&= \frac{\alpha}{P}\left(\frac{\alpha}{P}(\kappa_1 \frac{X F}{\sqrt{D}} + \kappa_\star Z)^T (\kappa_1 \frac{X F} {\sqrt{D}} + \kappa_\star Z) + \lambda \mathbb{I}\right)^{-1} \left(\kappa_1 \frac{X F} {\sqrt{D}} + \kappa_\star Z\right)^T X w_T \\
		&= \frac{\alpha}{P}\left(\frac{\alpha}{P}(\kappa_1 \frac{X F}{\sqrt{D}} + \kappa_\star Z)^T (\kappa_1 \frac{X F} {\sqrt{D}} + \kappa_\star Z) + \lambda \mathbb{I}\right)^{-1} \left(\kappa_1 \frac{X F} {\sqrt{D}} + \kappa_\star Z\right)^T X w_T \\
		&= \frac{\alpha}{P}\left(\frac{\alpha}{P}\left(\kappa_1^2 \frac{F^T X^T X F}{D} + \kappa_\star^2 Z^T Z + \kappa_1 \kappa_\star \left(\frac{Z ^T X F + F^T X Z} {\sqrt{D}}\right)\right) + \lambda \mathbb{I}\right)^{-1} \\
		&\times\left(\kappa_1 \frac{F^T X^T X w_T} {\sqrt{D}} + \kappa_\star Z^T X w_T\right)
	\end{split}
\end{equation}
By squaring this expression we can get the norm $q_d\, N=\lVert w \rVert^2$ of the OLS estimator. We would like to understand the behavior of this quantity when $P/N\to\infty$ and when $D/N=\alpha_D=\mathcal{O}(1)$ is varied. We can thus perform an annealed average over the dataset, averaging out $X$, $Z$, and $w_T$. 
Since we are going to square the expression, we can take advantage of:
\begin{equation}
	\mathbb{E} \frac{X^T X}{P} = \mathbb{E} \frac{Z^T Z}{P} = \mathbb{I}
\end{equation}
and defining $\Omega=F^T F / D$ we can simplify the expression as:
\begin{equation*}
	\mathbb{E}\lVert w \rVert^2 = \left(\kappa_1 \frac{w_T^T F} {\sqrt{D}} + \alpha \kappa_\star w_T^T\frac{X^T Z}{P} \right)\left(\left(\alpha \left(\kappa_1^2 \Omega + \kappa_\star^2 \mathbb{I}\right) + \lambda \mathbb{I}\right)^{-1}\right)^T \times
\end{equation*}
\begin{equation}
	\times\left(\alpha\left(\kappa_1^2 \Omega + \kappa_\star^2 \mathbb{I}\right) + \lambda \mathbb{I}\right)^{-1} \left(\kappa_1 \frac{F^T w_T} {\sqrt{D}} + \alpha \kappa_\star \frac{Z^T X}{P} w_T\right)
\end{equation}
If we now move to the eigenbasis of $\Omega=V \Lambda X^T$, where we also have that $F / \sqrt{D} = U \sqrt{\Lambda} V^T$, we can write this expression as a trace over the eigenvalues in $\Lambda$:
\begin{equation}
	q_d = \mathbb{E}\frac{\lVert w \rVert^2}{N} = \frac{\sigma^2_{w_T}}{N} \sum_{i=1}^N \frac{(\kappa_1^2 \rho_i + \alpha \kappa_\star^2)}{\left((\kappa_1^2 \rho_i + \kappa_\star^2) + \lambda \right)^2}
\end{equation}
Similarly, one can get an expression for the average overlap:
\begin{equation}
	Q_d = \mathbb{E}\frac{ w^T (\kappa_1^2\Omega + \alpha\kappa_\star^2) w }{N} = \frac{\sigma^2_{w_T}}{N} \sum_{i=1}^N \frac{(\kappa_1^2 \rho_i + \kappa_\star^2)^2}{\left((\kappa_1^2 \rho_i + \kappa_\star^2) + \lambda \right)^2} \label{eq::Q_d}
\end{equation}
So if we now focus on the ratio $q_d/Q_d$, which determines the fluctuations of the MD above $MD=1$, we can see the impact of the spectrum of $\Omega$, which follows a Marchenko-Pastur law with parameter $1/\alpha_D$. When $\alpha_D>1$ the spectrum is continuous and strictly positive, with a minimum eigenvalue $\rho_- = (1-\sqrt{1/\alpha_D})^2$. At $\alpha_D=1$ the spectrum touches the origin, and then at smaller values of $\alpha_D$ (in the overparameterized regime of the RFM) the distribution splits into a delta in $0$ with weight $1-\alpha_D$ and a continuous component with increasing left extremum $\rho_- = (1-\sqrt{1/\alpha_D})^2$ and weight $\alpha_D$.
Because of the additional $\rho_i$ in the numerator of expression \eqref{eq::Q_d}, when the eigenvalues of $\Omega$ approach zero they have a larger effect in $q_d$, therefore the MD reaches a peak. 
\begin{figure}[H]
	\captionsetup{justification=raggedright, singlelinecheck=false}
	\includegraphics[width=0.5\textwidth]{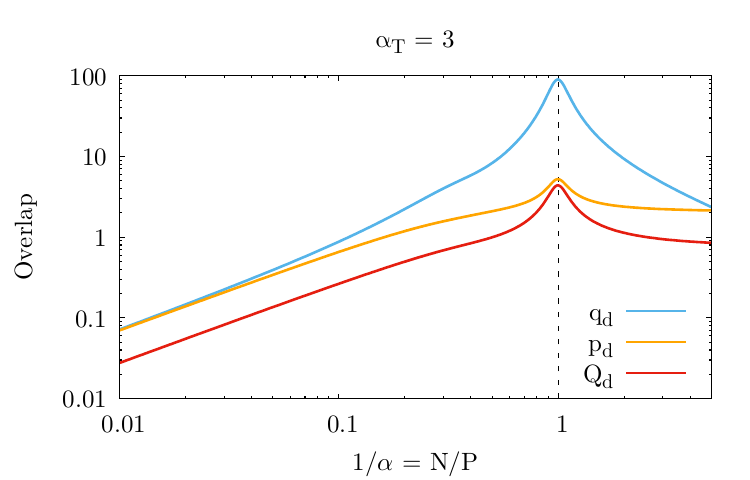}
	\includegraphics[width=0.5\textwidth]{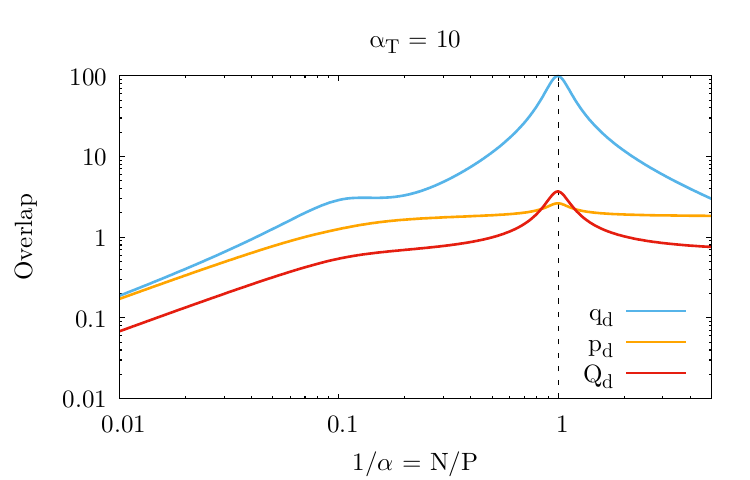}
	\center\includegraphics[width=0.5\textwidth]{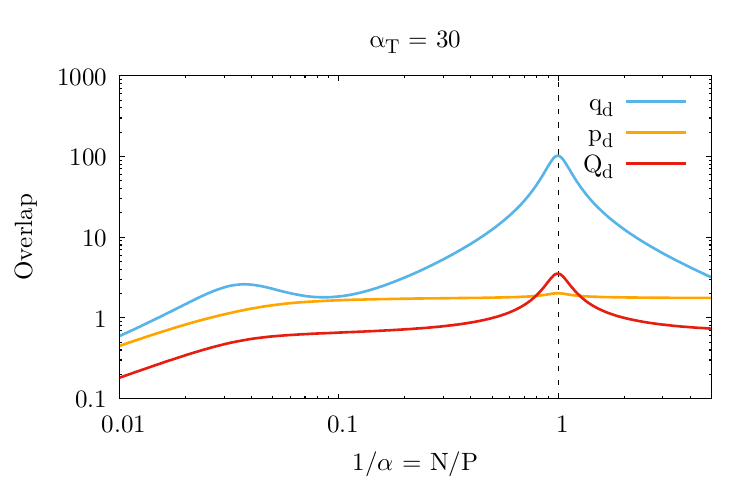}
	\caption{Plots of the overlaps $q_d$, $p_d$ and $Q_d = \kappa_1^2 p_d + \kappa_\star^2 q_d$ as a function of $1/\alpha$ for $\alpha_T = 3, 10, 30$. In all plots the loss used is MSE, the regularization is $\lambda = 10^{-4}$ and $\sigma = \tanh$.}
	\label{fig:Overlaps_anal}
\end{figure}
The same relationship between the parameters holds also at finite $\alpha$ and for a generic loss. The corresponding saddle-point equations read:
\begin{equation}
	q_d = \frac{1}{N} \sum_{i=1}^N \frac{((\hat{m}^2 + \hat{q})\kappa_1^2 \rho_i + \hat{q} \kappa_\star^2)}{\left(\delta \hat{q}(\kappa_1^2 \rho_i + \kappa_\star^2) + \lambda \right)^2}
\end{equation}
\begin{equation}
	Q_d = \frac{1}{N} \sum_{i=1}^N \frac{((\hat{m}^2 + \hat{q})\kappa_1^2 \rho_i + \hat{q} \kappa_\star^2)(\kappa_1^2 \rho_i + \kappa_\star^2)}{\left(\delta \hat{q}(\kappa_1^2 \rho_i + \kappa_\star^2) + \lambda \right)^2}
\end{equation}
where the loss function and the number of constraints determine the value of the conjugate parameters $\hat{m}, \hat{q}, \delta \hat{q}$, but the presence of an additional $(\kappa_1^2 \rho_i + \kappa_\star^2)$ in the numerator of $Q_d$ induces the same behavior of the MD around $\alpha_D=D/N=1$, independent of the specific setting.

In Fig.~\ref{fig:Overlaps_anal} we show the plot of the overlaps $q_d$, $p_d$ and $Q_d$ that confirms the intuitions showed above.

\section{Self-averaging property of the MD}
The MD of the RFM at initialization demonstrates the self-averaging properties, e.g. for the case of i.i.d. gaussian weights that were projected on the space orthogonal to the $\mathbb{I}_D$ vector.

\begin{figure}[H]
	\captionsetup{justification=raggedright,singlelinecheck=false}
	\includegraphics[scale=0.46]{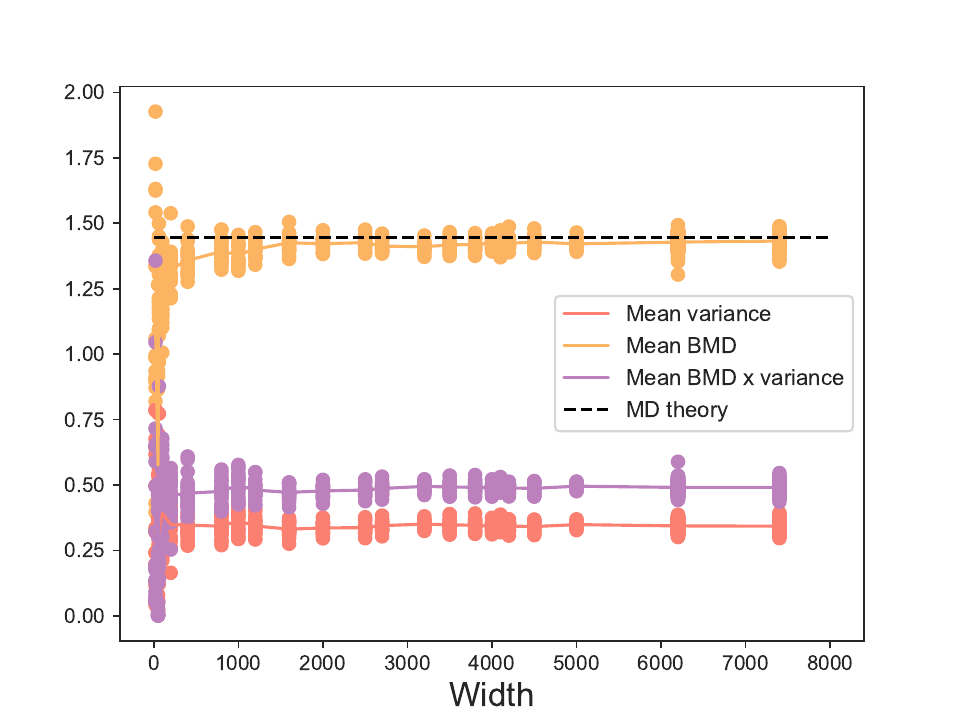}
	\includegraphics[scale=0.46]{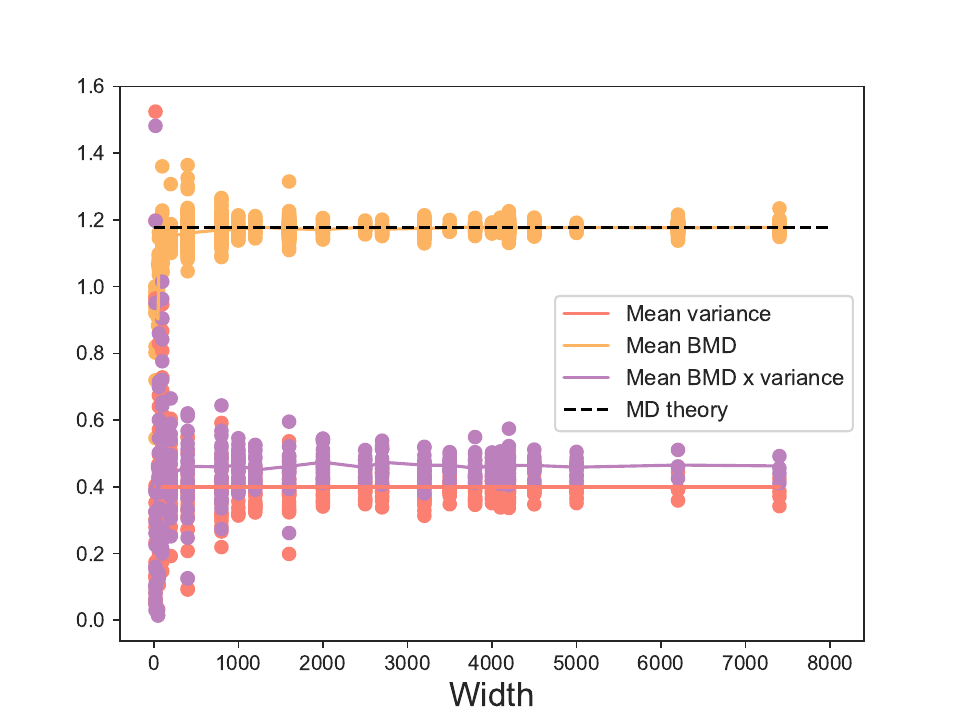}
	\caption{Correspondence of the empirical evidence of the self-averaging of the BMD computed for the RFM initialized with  the gaussian weights (projected orthogonally to the $\mathbb{I}_D$ vector) and theoretical prediction of the asymptotic BMD value in the limit of the input size $D$ and the hidden layer size $N$. (Left) RFM with leaky ReLU activation, (Right) RFM with tanh activation. The resulting plot represents an average over 70 different runs of the experiment.} 
	\label{fig:self_averaging}
\end{figure}

\subsection{Self-averaging of the MD for the trained models}\label{appC}

In the Table \ref{table:MDvariances} below we can observe the phenomenon of the concentration of the MD in the trained deep learning models as compared to the MD of the models at initialization.

\begin{table}[!h]
	\centering
	\begin{tabular}{*3c}
		Model   & Var(MD) trained model & Var(MD) untrained model\\
		\hline
		\text{ResNet20}   & 0.02503&	2.78465\\
		
		\text{ResNet32}   &0.0393&	21.49987\\
		
		\text{ResNet44}   &0.05377&	138.04827\\
		
		\text{ResNet56}   &0.0449&	792.22086\\
		
		\text{mobilenetv2 x0 5}   &0.07485&	318.17157\\
		
		\text{mobilenetv2 x0 75}   &0.05270&	271.84772\\
		
		\text{mobilenetv2 x1 0}   &0.04614&	81.66546\\
		
		\text{mobilenetv2 x1 4}  &0.03148&	128.32706\\
		
		\text{shufflenetv2 x0 5}   &0.02638&	146.33676\\
		
		\text{shufflenetv2 x1 0}   &0.0266&	34.79431\\
		
		\text{shufflenetv2 x1 5}   &0.02843&	27.3121\\
		
		\text{shufflenetv2 x2 0}   &0.03713&	24.21249\\
		
		\text{repvgg a0}   &0.03894&	104.79538\\
		
		\text{repvgg a1}  &0.04875&	109.59382\\
		
		\text{repvgg a2}   &0.05133&	136.44133\\
		
		\text{repvgg a0}   &3.44696	&74.58045\\
		
		\text{repvgg a1}  &4.27395	&83.92942\\
		
		\text{repvgg a2}   &2.66022	&114.86224\\
		
		\bottomrule
	\end{tabular}
	\caption{MD empirical variances for randomly initialized and trained on cifar-10 dataset deep models estimated over 30 seeds}
	\label{table:MDvariances}
\end{table}

\section{BMD and Data Normalization}

\label{sec:data_range}

\begin{figure}[H]
	\captionsetup{justification=raggedright, singlelinecheck=false}
	\centering \includegraphics[scale=0.75]{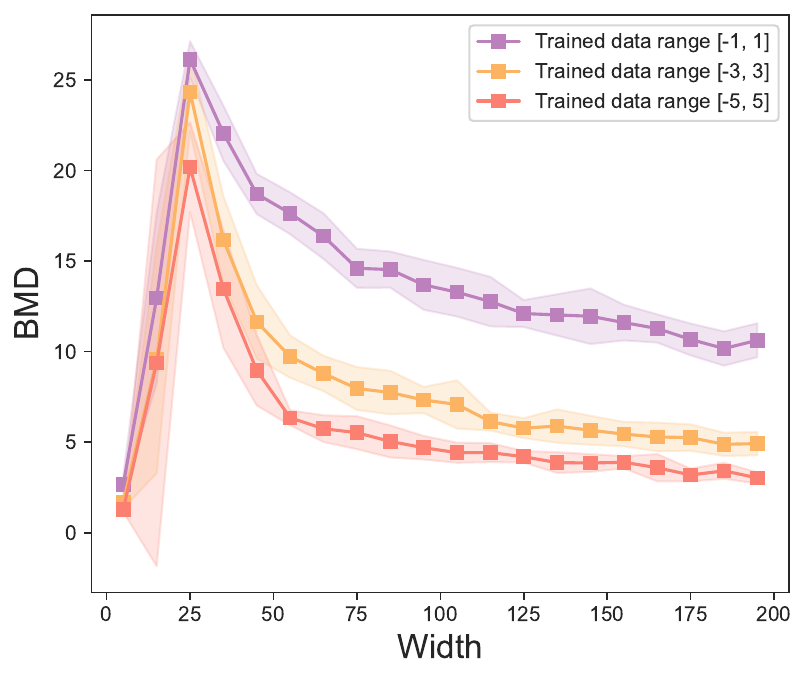}
	\caption{BMDs of the three RFMs trained on the MNIST dataset with 10 classes on 200 train samples with 0\% label noise. The RFMs were trained on the datapoints normalized to lie within the intervals [-5, 5] (red line), [-3, 3] (orange line) and [-1, 1] (violet line). The resulting plot represents an average (and standard deviations) obtained repeating 12 different times the experiment.}
	\label{fig:diff_ranges}
\end{figure}

In the Fig. \ref{fig:diff_ranges}, we repeat the BMD calculations for RFMs trained with different ranges used for normalization. As can be seen in the figure the choice of the input data normalization does not affect the BMD pattern, but only its absolute values in this setting.

\section{Effect of the label corruption on the train error}

\begin{figure}[H]
	\captionsetup{justification=raggedright, singlelinecheck=false}
	\centering \includegraphics[scale=0.75]{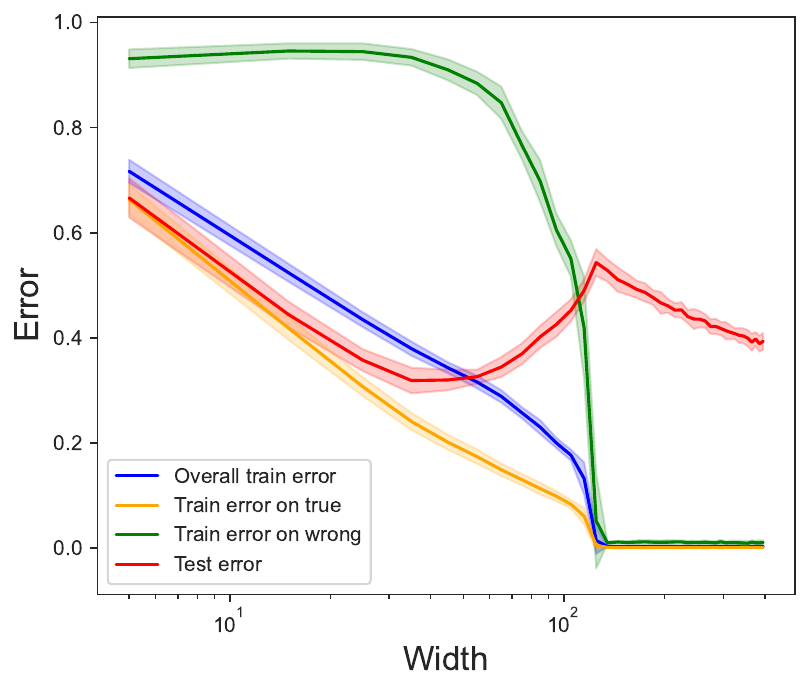}
	\caption{Train error curves of the RFM trained on the MNIST dataset with 10 classes on 1000 train samples with  20 \% label noise and estimated on 1000 test samples. The plot demonstrates that better performance of the model on the test data rather than on the train data for smaller widths can be explained by a disproportionally high error on the train examples with the corrupted (wrong) labels (green line), which therefore leads to the higher overall train error (blue line), while tested only on the data with non-corrupted labels the train error (yellow line) is comparable to the test error (red line)}
	\label{fig:wrong_labels}
\end{figure}

In the Fig. \ref{fig:wrong_labels} we demonstrate the effect of the label corruption of the train data on the train error, which is allowing to explain the difference between the test and train errors for smaller model widths.

\end{document}